\RequirePackage[svgnames,table]{xcolor}
\documentclass[11pt,letterpaper]{yalearxiv}

\usepackage{bm}
\usepackage{xspace}
\usepackage{adjustbox}
\usepackage{multicol}
\usepackage{natbib}
\usepackage{fontawesome5}

\setcitestyle{square}

\makeatletter
\def\munderbar#1{\underline{\sbox\tw@{$#1$}\dp\tw@\z@\box\tw@}}
\makeatother

\AddToHook{cmd/appendix/before}{%
  \setcounter{axiom}{0}%
}

\newcommand{\be}{\begin{equation}}
\newcommand{\ee}{\end{equation}}
\newcommand{\bea}{\begin{equation*}\begin{aligned}}
\newcommand{\eea}{\end{aligned}\end{equation*}}

\newcommand{\projectname}{\textsc{QATFactory}\xspace}
\definecolor{qadrow}{RGB}{234,242,252}

\title{\projectname{}: A Versatile, Deployment-Aligned Framework for Quantization-aware Training and Distillation of LLMs}
\runningtitle{\textsc{QATFactory}: A Versatile, Deployment-Aligned Framework for Quantization-aware Training and Distillation of LLMs}

\author{%
\textbf{Weili Xu\textsuperscript{1,2,*}},\
\textbf{Jisen Li\textsuperscript{1,*}},\
\textbf{Yuqing Jian\textsuperscript{1,*}},\
\textbf{Chenxi Li\textsuperscript{1}},\
\textbf{Zhizhou Sha\textsuperscript{1,3}},\
\textbf{Yifan Yu\textsuperscript{1,2}}\\
\vspace{2pt}

\textbf{Qingyang Wu\textsuperscript{1}},\
\textbf{Chenfeng Xu\textsuperscript{1,3}},\
\textbf{Zhongzhu Zhou\textsuperscript{1}},\
\textbf{Tianyi Zhang\textsuperscript{1}},\
\textbf{Ben Athiwaratkun\textsuperscript{1}}\\
\vspace{6pt}

\small
\textsuperscript{1}Together AI \qquad
\textsuperscript{2}University of Illinois Urbana-Champaign \qquad
\textsuperscript{3}The University of Texas at Austin\\
\vspace{5pt}

\small\textsuperscript{*}Equal contribution\\
\vspace{5pt}

\small\faGithub~\textbf{Source Code:}~
\href{https://github.com/QATFactory/QATFactory}
{\texttt{github.com/QATFactory/QATFactory}}%
}

\hypersetup{
  colorlinks=true,
  linkcolor=blue!50!black,
  citecolor=blue!50!black,
  urlcolor=blue!50!black
}
\begin{document}

\begin{abstract}
\vspace{-1mm}
{\centering\section*{Abstract}}
Large language model (LLM) inference is increasingly moving toward lower precision to realize the throughput of hardware accelerators, but aggressive post-training quantization (PTQ) can degrade model quality. We present QATFactory, an open-source framework for deployment-aligned quantization-aware distillation (QAD) and reinforcement learning (QARL). QATFactory simulates deployment-time quantization while performing matrix multiplications in BF16, allowing models to adapt to quantization noise without requiring training hardware that natively supports the target format; for example, it supports NVFP4 training on H100 GPUs, which lack FP4 Tensor Cores. The framework supports NVFP4, MXFP4, and llama.cpp's Q4\_K format; dense and mixture-of-experts models; and both full-parameter and LoRA-based training. It exports checkpoints directly to vLLM and llama.cpp without an additional lossy conversion step or added inference overhead. With QATFactory, we conduct extensive experiments on models ranging from 8B to 230B parameters and evaluate exported checkpoints in production inference engines. Across models and formats, QAD consistently improves deployed-model quality over strong PTQ baselines. On Qwen3.5-9B, QAD achieves average benchmark accuracies of 68.9\% under NVFP4 and 66.0\% under MXFP4, outperforming the best PTQ results of 65.4\% and 56.4\%, respectively. Through our experiments, we found that although both FP4 formats quantize weights and activations at deployment, the best training strategy is format-dependent: NVFP4 generally performs better when only weights are quantized during training, whereas MXFP4 benefits from quantizing both weights and activations. For QAD of Qwen3.5-9B, rank-16 LoRA reduces GPU memory usage by \(2.9\times\), but increasing the rank does not close the quality gap with full-parameter QAD. At a fixed training token budget, training on fewer 32K sequences improves average accuracy by 1.9 points over training on more 4K sequences. Beyond distillation, NVFP4 QARL improves end-to-end training throughput by \(1.23\times\) and average accuracy by 2.7 points over BF16 reinforcement learning followed by PTQ across five math benchmarks. We release the complete QATFactory training code and the resulting low-precision checkpoints.
\end{abstract}

\maketitle

\begin{figure}[h]
\centering
\includegraphics[width=\linewidth]{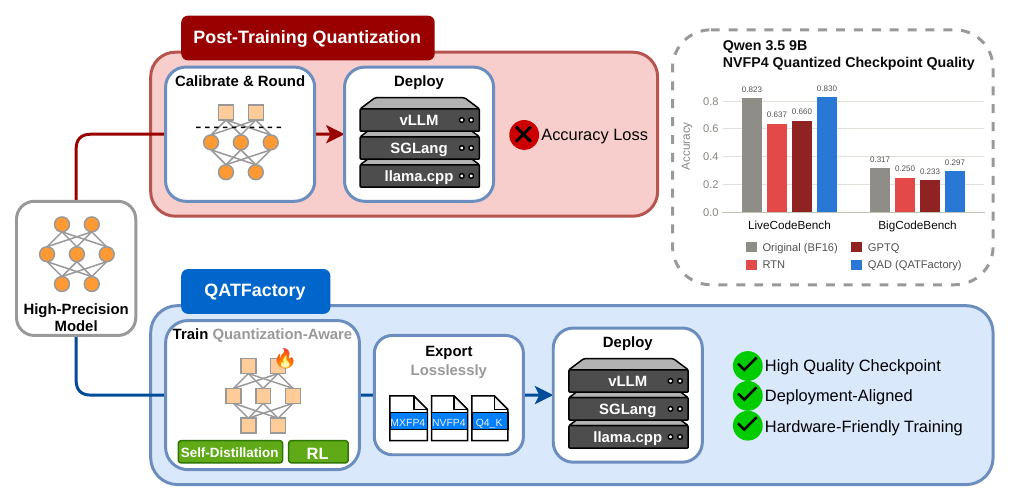}
\caption{Comparison of post-training quantization (PTQ) and quantization-aware training (QAT) in \projectname{}. PTQ directly quantizes a pretrained model and can lead to accuracy degradation. QATFactory instead trains the model with quantization in the loop using self-distillation or reinforcement learning, then exports deployment-ready low-precision checkpoints. On Qwen3.5-9B, QATFactory produces higher-quality NVFP4 models than RTN and GPTQ on both LiveCodeBench and BigCodeBench.
}
\label{fig:teaser}
\end{figure}

\section{Introduction}
\label{sec:intro}

Every new accelerator generation is making a more aggressive bet: useful model capacity will be served at progressively lower numerical precision. This bet is driven both by the economics of inference, which increasingly dominates the lifetime cost of deploying large language models (LLMs), and by the architecture of the hardware itself. On NVIDIA's Blackwell generation, a single B200 GPU delivers peak Tensor Core throughput of 4.5~PFLOPS in BF16, 9~PFLOPS in FP8, and 18~PFLOPS in FP4~\citep{nvidia2024blackwell}.\footnote{Per-GPU datasheet figures with structured sparsity; dense throughput is half of each figure (2.25, 4.5, and 9~PFLOPS, respectively). Each Blackwell package comprises two reticle-limited dies, but the 1:2:4 scaling across BF16, FP8, and FP4 holds under either convention.} Exploiting low-precision formats, particularly 4-bit formats such as NVFP4 and MXFP4~\citep{rouhani2023microscaling,nvidia2025nvfp4}, is therefore becoming a prerequisite for realizing the computational and economic value of modern accelerators. For inference providers, failing to exploit the low-precision formats increasingly means leaving a substantial fraction of available throughput on the table.

Yet the economic incentive to reduce precision creates a fundamental tension: the formats that offer the greatest hardware efficiency are also the least tolerant of numerical error. Each accelerator generation places a larger fraction of its peak throughput behind increasingly aggressive representations, even as the reasoning and agentic models being deployed become more sensitive to small perturbations in their computations. NVIDIA's next-generation Rubin architecture, for example, introduces a programmable lookup-table Tensor Core format in which each weight stores a 3-bit index into a shared, non-uniform codebook~\citep{semianalysis2026rubin}. However, whether such emerging 3-bit formats can preserve usable model quality will depend not only on codebook design and calibration data, but increasingly on whether models are explicitly trained to withstand their deployment numerics~\citep{semianalysis2026rubin}. The central question is therefore shifting from \emph{how to quantize a model after it has been trained} to \emph{how to train models whose behavior survives the numerical constraints of the hardware on which they will run}.

Today's dominant answer remains post-training quantization (PTQ): train a high-precision model, collect a modest calibration set, and quantize the resulting checkpoint without updating its parameters. This separation works well at 8 bits and often remains adequate for weight-only 4-bit quantization. It breaks down, however, at W4A4---the regime that actually unlocks FP4 tensor-core throughput. Unlike W4A16, where only the weights incur a static perturbation while activations remain expressive, W4A4 injects error into both operands of every matrix multiplication. FP4 provides little room to absorb this error: an E2M1 value occupies one of only 15 representable points over a coarse range before scaling, making substantial rounding and clipping unavoidable. These perturbations are especially damaging for modern reasoning and agentic models. Such models generate thousands of autoregressive tokens, repeatedly encounter high-entropy branching decisions, and condition every subsequent prediction on their earlier outputs~\citep{liu2025quantization,lotfi2026quantized,marcuzzi2025bias}.


Quantization-aware training (QAT) addresses the aforementioned problem by inserting quantization into the training forward pass, allowing the model parameters to adapt to the induced numerical perturbations. Quantization-aware distillation (QAD) further stabilizes this process by using the original high-precision model to supervise its quantized counterpart~\citep{jacob2018quantization,hinton2015distilling,liu2024llm,kim2019qkd,xin2026quantization}. Yet for modern deployment formats, the primary obstacle is no longer the absence of a training objective. It is the mismatch between the numerical semantics used during training and those implemented by production inference engines. Existing open-source QAT pipelines~\citep{torchao2024,liu2024llm} largely target generic integer quantization schemes rather than the floating-point and engine-native formats used in practice. Their fake quantizers may differ from inference kernels in representable values, clipping behavior, block structure, rounding, or scaling hierarchy. As a result, a checkpoint can train successfully yet fail to load in the target engine. Support for emerging formats such as NVFP4 and MXFP4, as well as engine-native formats such as Q4\_K, remains underexplored.

We introduce \projectname, a fully open framework for engine-aligned QAT and QAD across dense and mixture-of-experts (MoE) models~\citep{shazeer2017outrageously}. \projectname supports NVFP4, MXFP4, and llama.cpp's Q4\_K format~\citep{gerganov2023llamacpp}, together with both full-parameter training and parameter-efficient QAT through low-rank adaptation (LoRA)~\citep{hu2022lora}. Its central design principle is simple: \emph{treat the inference engine as the numerical specification}. For each target format, an engine-aligned fake quantizer reproduces its exact representable-value set, numerical range, block organization, rounding behavior, and scaling rules, including hierarchical two-level scaling, while retaining high-precision linear algebra during training. This separation provides deployment-faithful numerics without requiring hardware that natively executes the target format. The resulting checkpoints are directly consumable by production inference engines, including vLLM, SGLang, and llama.cpp~\citep{vllm,sglang,gerganov2023llamacpp}, without additional operators or runtime overhead. Moreover, because \projectname emulates the target arithmetic rather than executing it natively, it enables NVFP4 QAT on widely available H100 GPUs, despite their lack of NVFP4 Tensor Cores~\citep{nvidia_cuda_12_8}. The framework thus decouples access to low-precision training from access to the newest generation accelerator.

Beyond producing deployable low-precision checkpoints, \projectname provides a controlled experimental platform for studying which aspects of inference numerics and training design matter for low-precision adaptation. Holding the training infrastructure fixed while varying quantization formats, activation precision, adaptation strategies, sequence lengths, data domains, and training objectives reveals four findings: \textbf{(1)} faithfully reproducing every inference-time quantization operation during training is not universally optimal: MXFP4 benefits from quantizing both weights and activations during training, whereas NVFP4 generally produces higher-quality checkpoints with weight-only quantization, exposing a format-dependent tradeoff between deployment fidelity and optimization stability. \textbf{(2)} LoRA-based parameter-efficient QAD substantially reduces training memory, but underperforms full-parameter QAD in our experiments, and increasing the adapter rank alone does not close the quality gap. \textbf{(3)} under a fixed training-token budget, training on fewer long sequences generalizes better overall than training on more short sequences, and QAD improvements transfer across domains, with particularly strong transfer from code training to mathematical reasoning. \textbf{(4)} quantization-aware reinforcement learning improves both efficiency and deployed-model quality: NVFP4 QARL accelerates end-to-end RL training by $1.23\times$ while improving mean reasoning accuracy by 2.7 percentage points over BF16 RL followed by post-training quantization.


\noindent
\textbf{Contributions.}
\begin{itemize}[topsep=0pt]
\item \textbf{An engine-aligned QAT/QAD framework.}
We introduce an open framework spanning NVFP4, MXFP4, and Q4\_K; dense and MoE architectures; and full-parameter and LoRA-based training. Exported checkpoints load directly into vLLM, SGLang, and llama.cpp without custom operators or runtime overhead.

\item \textbf{Hardware-independent low-precision training.}
By emulating target-format numerics while retaining high-precision computation, \projectname enables NVFP4 QAT on widely available H100 GPUs and removes the requirement that training hardware natively support the deployment format.

\item \textbf{New insights into low-precision optimization.}
We identify a format-dependent asymmetry in activation quantization,
characterize the memory--quality trade-off of parameter-efficient QAD,
and show that increasing the LoRA adapter rank does not close the gap
to full-parameter QAD.

\item \textbf{Broad empirical validation and open release.}
Across model families, architectural classes, and scales, QAD-trained checkpoints consistently outperform strong PTQ baselines and approach BF16 quality when evaluated directly in production inference engines. We release the framework, training recipes, evaluation tools, and trained low-precision models.

\end{itemize}

\section{Related Work}

\paragraph{Post-training quantization for LLMs.}
Post-training quantization (PTQ) is the dominant approach for reducing the memory and inference cost of large language models without retraining.
GPTQ~\citep{frantar2022gptq} uses approximate second-order information to minimize weight reconstruction error during quantization, while SmoothQuant~\citep{xiao2023smoothquant} redistributes quantization difficulty between weights and activations to enable accurate low-precision inference.
AWQ~\citep{lin2024awq} further exploits activation statistics to preserve salient weight channels under low-bit weight quantization.
Although these methods achieve strong accuracy--efficiency trade-offs, aggressive quantization becomes increasingly challenging when both weights and activations are reduced to very low precision.
Recent studies show that quantization can disproportionately affect reasoning behavior, generation trajectories, and other model properties~\citep{liu2025quantization,lotfi2026quantized,marcuzzi2025bias}.
These observations motivate training-time adaptation to the numerical perturbations introduced by deployment.

\paragraph{Quantization-aware training and distillation.}
Quantization-aware training (QAT) exposes a model to simulated quantization during optimization, allowing its parameters to adapt to low-precision numerical constraints.
Classical QAT methods use fake quantization together with straight-through estimators to optimize continuous latent parameters under discretized forward computations~\citep{jacob2018quantization}.
Subsequent work improves the optimization of quantized networks by learning activation clipping thresholds, as in PACT~\citep{choi2018pact}, and quantization step sizes, as in LSQ~\citep{esser2019learned}.
Quantization Noise~\citep{fan2020training} further studies training under simulated quantization perturbations for extreme model compression.

Early work on Transformer quantization extends low-precision training and adaptation to pretrained language models.
Q-BERT~\citep{shen2020q} studies ultra-low-precision quantization of BERT using Hessian information, while TernaryBERT~\citep{zhang2020ternarybert} combines ultra-low-bit quantization with knowledge distillation to recover model quality.
I-BERT~\citep{kim2021bert} further demonstrates integer-only inference for Transformer models by quantizing both linear and nonlinear operations.
For LLMs, LLM-QAT~\citep{liu2024llm} extends this paradigm through data-free distillation and studies joint quantization of weights, activations, and KV caches.
BitDistiller~\citep{du2024bitdistiller} further demonstrates that self-distillation can substantially improve the robustness of extremely low-bit LLMs.
More recently, \cite{xin2026quantization} study quantization-aware distillation (QAD) specifically for NVFP4 and show that a high-precision teacher can recover substantial accuracy lost during FP4 deployment.

\paragraph{Efficient and scalable low-precision training.}
The cost of full-parameter QAT has motivated work on more efficient optimization and parameter-efficient adaptation.
EfficientQAT~\citep{chen2025efficientqat} develops efficient training procedures for low-bit LLM adaptation, while QA-LoRA~\citep{xu2024qa} combines quantization awareness with low-rank adaptation to reduce the trainable parameter and memory footprint.
LR-QAT~\citep{bondarenko2024low} introduces a memory-efficient low-rank QAT formulation for LLMs that produces fully quantized models without additional inference overhead, while L4Q~\citep{jeon2025l4q} integrates low-rank adaptation with QAT for parameter-efficient quantization-aware fine-tuning.
SketchTune~\citep{zhang2025sketch} explores a complementary LUT-based approach: each weight is represented by an index into a compact lookup table (LUT) of shared values, and adaptation directly optimizes these LUT entries while retaining the index assignments.
Complementary studies characterize the scaling behavior of QAT itself.
ParetoQ~\citep{liu2026paretoq} investigates optimization regimes across extremely low-bit quantization settings, and \citet{chen2025scaling} study how QAT performance varies with model scale, training data, and quantization configuration.

\paragraph{FP4 and deployment-native low-precision training.}
The emergence of native FP4 hardware has motivated increasing interest in training models directly under FP4 numerical constraints.
NVFP4 combines E2M1 values with fine-grained block scaling and a higher-level tensor scale to improve the accuracy of FP4 computation~\citep{nvidia2025nvfp4}, while microscaling formats such as MXFP4 use shared block-level scales to enable efficient low-precision arithmetic~\citep{rouhani2023microscaling}.
Recent work has demonstrated large-scale pretraining with NVFP4~\citep{abecassis2025pretraining} and MXFP4~\citep{cim2026pretraining}, while Full-Stack FP4~\citep{ding2026full} extends FP4 computation across multiple components of the training stack.

\paragraph{Low-precision reinforcement learning.}
Recent work has also explored low-precision computation in reinforcement learning for LLM post-training.
QuRL~\citep{li2026qurl} studies low-precision reinforcement learning for efficient reasoning, while QeRL~\citep{huang2026qerl} combines quantization with reinforcement learning to improve training efficiency and investigates quantization as more than a purely deployment-time compression mechanism.
QaRL~\citep{gu2026qarl} studies rollout-aligned quantization-aware reinforcement learning and addresses the numerical mismatch between low-precision rollout generation and policy optimization.
More recently, Miles~\citep{chen2026miles} presents a production-level post-training system that includes low-precision training and rollout workflows, with explicit attention to consistency between training-time quantization, weight synchronization, and low-precision inference.

\section{QATFactory}
\label{sec:qatfactory}

QATFactory is an open-source framework for adapting pretrained large language models (LLMs) to deployment-native quantization formats through quantization-aware training (QAT) or distillation. It simulates quantization and dequantization during training while performing all matrix multiplications in BF16. This design allows models to be trained for low-bit formats, such as NVFP4, MXFP4, and Q4\_K, without requiring specialized hardware. After training, QATFactory exports packed checkpoints that run directly in target inference engines without extra post-training quantization (PTQ).

The framework supports two training workflows: quantization-aware distillation (QAD) and quantization-aware reinforcement learning (QARL). QAD recovers accuracy by training a quantized model to match its frozen high-precision counterpart. QARL trains the quantized model with reward or verifier feedback while using low-precision inference engines for efficient rollout generation. Both workflows share the same quantization layers, format implementations, and export pipeline.

\subsection{Framework overview and design goals}
\label{sec:overview}

\begin{figure*}[t]
\centering
\includegraphics[width=\textwidth]{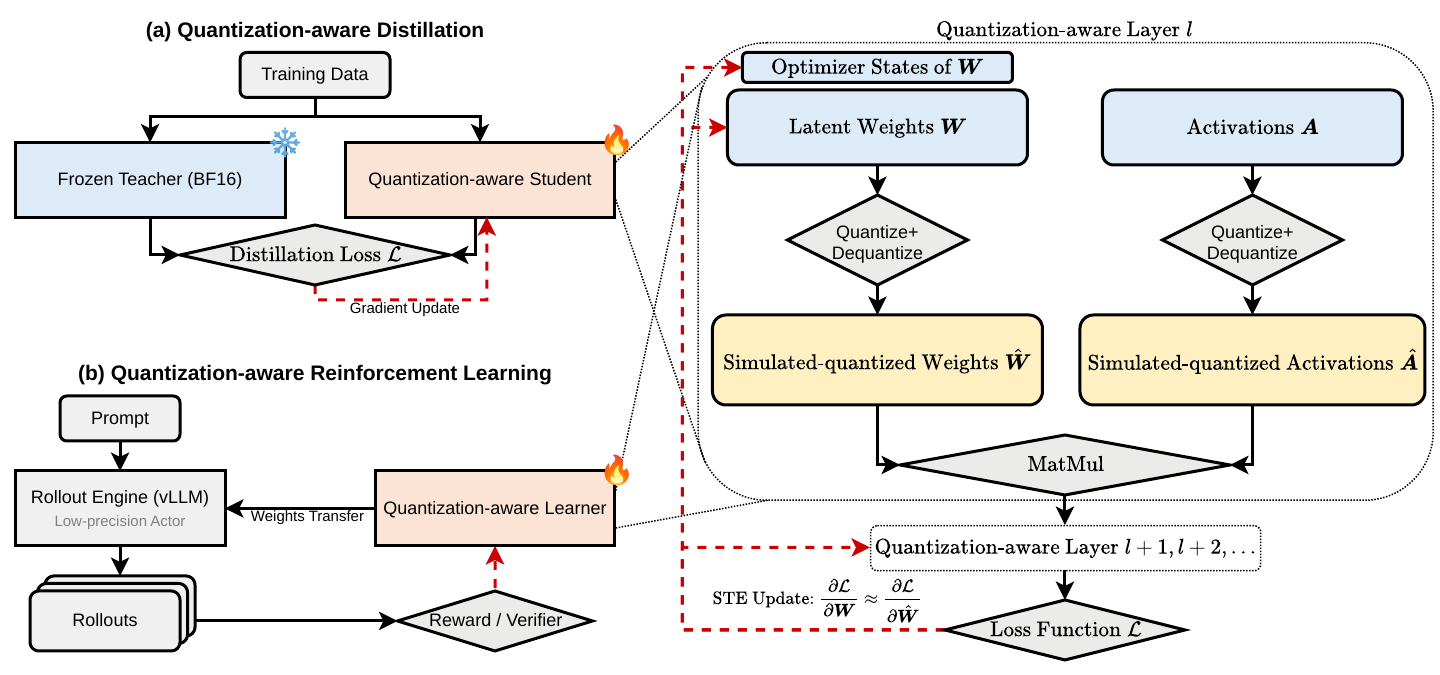}
\caption{Overview of QATFactory. \textbf{(a) Quantization-aware distillation (QAD):} a frozen high-precision teacher supervises a quantization-aware student initialized from the same pretrained checkpoint. \textbf{(b) Quantization-aware reinforcement learning (QARL):} a low-precision inference engine generates rollouts, reward functions or verifiers evaluate them, and updated weights are synchronized back to the rollout engine. \textbf{Right:} the shared internal structure of a quantization-aware layer. Latent weights and activations undergo simulated quantization in BF16, and parameter updates flow through a straight-through estimator (STE).}
\label{fig:overview}
\end{figure*}

\paragraph{Motivation and pain points.}
Applying quantization-aware training in practice faces two key challenges. First, existing QAT frameworks often focus on generic formats (such as integer W4A8~\citep{pytorch-qat}) that lack native support in modern inference engines like vLLM, SGLang, and \texttt{llama.cpp}~\citep{vllm,sglang,gerganov2023llamacpp}. Second, low-bit training may introduce steep hardware and memory barriers. Training natively in low precision may require specialized hardware with format-specific compute support, such as GPUs with native FP4 matrix multiplication capability. In addition, full-parameter distillation incurs high memory overhead because teacher weights, student latent weights, and optimizer states must all reside in GPU memory simultaneously.

\paragraph{Framework design.}
Figure~\ref{fig:overview} shows the design of QATFactory. Both QAD and QARL use deployment-aligned quantization-aware layers. During the forward pass, these layers apply a lossy quantize--dequantize transformation to reproduce inference-time numerical constraints. During the backward pass, a straight-through estimator (STE)~\citep{ste} updates the continuous latent parameters. QATFactory supports both full-parameter training and memory-efficient adaptation with LoRA~\citep{hu2022lora}. This design is guided by three goals:
\begin{enumerate}
\item \textbf{Faithfulness to deployment.} QATFactory supports widely used deployment formats, and its training process reflects the numerical precision and quantization format used during inference. This alignment exposes the model to deployment-time quantization error during training and allows direct checkpoint export without additional quantization or calibration (\S\ref{sec:fakequant}, \S\ref{sec:export}).
\item \textbf{Extensibility.} QATFactory can be easily extended to support new quantization formats and target inference engines. Each format implements a standard interface for quantize--dequantize operations and checkpoint packing, keeping training loops independent of format-specific logic (\S\ref{sec:fakequant}).
\item \textbf{Accessibility and scalability.} Because all matrix multiplications run in BF16, training does not depend on specialized low-bit compute kernels. FSDP~\citep{10.14778/3611540.3611569} enables scaling efficiently to large models, while LoRA provides a more memory-efficient path for adapting large models (\S\ref{sec:fakequant}, \S\ref{sec:qad}).
\end{enumerate}

\subsection{Deployment-aligned simulated quantization}
\label{sec:fakequant}

To faithfully adapt models without requiring low-bit training hardware, QATFactory replaces standard linear projections with \emph{quantization-aware layers} that simulate deployment-time numerical behavior. Each layer is parameterized by a \emph{deployment configuration}, which pairs a quantization format (e.g., NVFP4, MXFP4, or Q4\_K) with a target inference engine (e.g., vLLM, SGLang, or \texttt{llama.cpp}). This section explains how these layers simulate low-precision constraints during training and align with engine-specific semantics. It also describes the unified quantizer interface that supports multiple formats.

\subsubsection{Quantization-aware layer}
\label{sec:qalayer}

During deployment, inference engines store and compute with low-bit weights and activations natively, which can require hardware with native arithmetic support for formats such as NVFP4 and MXFP4. In contrast, QATFactory maintains weights and activations in high precision during training and applies \textit{simulated quantization} to reproduce values representable in the deployment format. This process exposes the model to quantization error while keeping all matrix multiplications in BF16.

Let \(\mathcal{Q}_{\kappa}\) and \(\mathcal{Q}^{-1}_{\kappa}\) denote the quantization and dequantization operators for deployment configuration \(\kappa\). Given a high-precision tensor \(\bm{X}\), the quantizer produces low-bit codes \(\bm{C}\) and format-specific metadata \(\bm{\theta}\):
\begin{equation}
  \mathcal{Q}_{\kappa}(\bm{X})
  \rightarrow
  (\bm{C},\bm{\theta}),
  \qquad
  \mathcal{Q}^{-1}_{\kappa}(\bm{C},\bm{\theta})
  \rightarrow
  \hat{\bm{X}}.
  \label{eq:quantize-dequantize}
\end{equation}
The metadata \(\bm{\theta}\) contains the scaling factors and zero points needed to dequantize the low-bit codes into full-precision floating-point values. The reconstructed tensor \(\hat{\bm{X}}\) remains in BF16, but its values are strictly restricted to the grid representable by the target low-bit format.

For a high-precision latent weight matrix \(\bm{W}\) and input activations \(\bm{A}\), the quantization-aware layer computes
\begin{equation}
  \hat{\bm{W}}
  =
  \mathcal{Q}^{-1}_{\kappa}
  \left(
    \mathcal{Q}_{\kappa}(\bm{W})
  \right),
  \qquad
  \hat{\bm{A}}
  =
  \mathcal{Q}^{-1}_{\kappa}
  \left(
    \mathcal{Q}_{\kappa}(\bm{A})
  \right),
  \qquad
  \bm{O}
  =
  \hat{\bm{W}}\hat{\bm{A}}.
  \label{eq:fakequant}
\end{equation}
Because the latent weights change after every optimization step, their quantized representations cannot be cached. QATFactory therefore recomputes them during each forward pass, immediately before matrix multiplication. The quantizer derives tensor- and block-level scale factors from the current weights according to the target format, typically using the maximum absolute value for symmetric quantization. Activations are likewise quantized on the fly. We implement these quantize--dequantize operations with efficient GPU kernels to limit their training-time overhead. The resulting fake-quantized tensors \(\hat{\bm{W}}\) and \(\hat{\bm{A}}\) exist only transiently in GPU memory and are discarded immediately after the matrix multiplication.

Because discrete quantization has zero derivative almost everywhere, QATFactory uses a straight-through estimator (STE) during the backward pass, treating the quantize--dequantize transformation as an identity mapping:
\begin{equation}
  \frac{\partial \hat{\bm{W}}}{\partial \bm{W}}
  \approx
  \bm{I},
  \qquad
  \frac{\partial \mathcal{L}}{\partial \bm{W}}
  \approx
  \frac{\partial \mathcal{L}}{\partial \bm{O}}
  \hat{\bm{A}}^{\mathsf{T}}.
  \label{eq:ste}
\end{equation}
This allows gradients to pass directly through the quantization operations and update the continuous latent weights and optimizer states in high precision.

\subsubsection{Quantizer interface}
\label{sec:quantizer-interface}

QATFactory places format-specific logic behind a unified quantizer interface. Each quantizer defines the quantize--dequantize behavior of a specific data format and exports the resulting low-precision weights in a layout compatible with the target inference engine. During training, it reproduces the numerical behavior of the target inference format by partitioning tensors into blocks, computing scales, mapping values to the representable grid, and reconstructing BF16 operands for matrix multiplication. During export, it losslessly packs the low-precision weights and metadata into the layout expected by the target inference engine. This keeps the quantization logic separate from the training logic, making the support of new quantization formats straightforward.

\subsubsection{Supported quantization formats}
\label{sec:formats}

We briefly introduce the characteristics of the primary quantization formats that QATFactory supports, including NVFP4, MXFP4, and Q4\_K. Table~\ref{tab:formats} provides an overview for these data formats.

\begin{table}[t]
  \centering
  \small
  \begin{tabular}{lccc}
    \toprule
    & \textbf{NVFP4} & \textbf{MXFP4} & \textbf{Q4\_K} \\
    \midrule
    Precision
      & W4A4
      & W4A4
      & W4A16 (Weight-only) \\
    Element format
      & E2M1 (FP4)
      & E2M1 (FP4)
      & Unsigned INT4 \\
    Block structure
      & Block size 16
      & Block size 32
      & 32 (in 256 super-block) \\
    Scale hierarchy
      & Block E4M3, Tensor FP32
      & Block E8M0
      & Block 6-bit, Super-block FP16 \\
    Symmetry
      & Symmetric
      & Symmetric
      & Asymmetric \\
    Effective bits / weight
      & \(4.50\) bits
      & \(4.25\) bits
      & \(4.50\) bits \\
    Target engines
      & vLLM, SGLang
      & vLLM, SGLang
      & \texttt{llama.cpp} \\
    \bottomrule
  \end{tabular}
  \caption{Quantization formats supported by QATFactory. Effective bits per weight include element codes and scaling metadata. W4A4 denotes 4-bit weights and activations; W4A16 denotes 4-bit weights with high-precision activations.}
  \label{tab:formats}
\end{table}

\paragraph{NVFP4.}
NVFP4~\citep{nvidia2025nvfp4} represents weights using the 4-bit floating-point format \textbf{E2M1}, which uses a 1-bit sign, a 2-bit exponent, and a 1-bit mantissa. This format represents exactly 16 discrete values:
\begin{equation}
  \mathcal{V}_{\mathrm{E2M1}} = \left\{ 0, \pm 0.5, \pm 1, \pm 1.5, \pm 2, \pm 3, \pm 4, \pm 6 \right\}.
  \label{eq:e2m1-values}
\end{equation}
NVFP4 employs a two-level scale hierarchy. Each block of 16 weights is represented by 16 E2M1 element codes and an 8-bit E4M3 block scale. A global tensor-level FP32 scale sets the overall dynamic range and normalizes the block scales:
\begin{equation}
  \mathcal{Q}_{\mathrm{NVFP4}}(\bm{W})
  \rightarrow
  \left(
    \bm{C}_{\mathrm{E2M1}},
    \bm{s}_{\mathrm{E4M3}},
    s_{\mathrm{FP32}}
  \right).
  \label{eq:nvfp4}
\end{equation}
For an NVFP4 weight $i$ in block $b$, dequantization multiplies the element code by both the block scale and the tensor scale:
\begin{equation}
  \hat{W}_{i} = s_{\mathrm{FP32}} \cdot s_{\mathrm{E4M3}, b} \cdot C_{i},
  \label{eq:nvfp4-dequant}
\end{equation}
where $C_i \in \mathcal{V}_{\mathrm{E2M1}}$.

During NVFP4 inference, BF16 or FP32 activations are quantized to FP4 immediately before each matrix multiplication using a static tensor-level FP32 scale and per-block E4M3 scales computed on the fly. PTQ typically obtains the tensor-level FP32 activation scale for each layer through offline calibration, whereas QATFactory derives it from the running absolute maximum ($\operatorname{amax}$) of each layer's input activations during training and exports it with the checkpoint.

\paragraph{MXFP4.}
MXFP4~\citep{mxfp4} also uses E2M1 element codes, grouping weights into blocks of 32 values under a shared 8-bit scale factor $s_{\mathrm{E8M0}, b} \in \text{E8M0}$:
\begin{equation}
  \mathcal{Q}_{\mathrm{MXFP4}}(\bm{W})
  \rightarrow
  \left(
    \bm{C}_{\mathrm{E2M1}},
    \bm{s}_{\mathrm{E8M0}}
  \right).
  \label{eq:mxfp4}
\end{equation}
Unlike NVFP4, MXFP4 has no tensor-level scale. Because E8M0 contains 8 exponent bits and no mantissa, each block scale is an exact power of two ($2^{p_b}$ for integer $p_b \in [-127, 128]$). Dequantizing an MXFP4 weight requires only an exponent shift:
\begin{equation}
  \hat{W}_{i} = 2^{p_b} \cdot C_{i}.
  \label{eq:mxfp4-dequant}
\end{equation}
Both NVFP4 and MXFP4 are W4A4 formats, quantizing weights and activations to FP4 during deployment. MXFP4 uses fewer effective bits per weight than NVFP4 (4.25 versus 4.50) by sharing each scale across 32 values rather than 16. Compared with NVFP4, MXFP4 uses larger groups and coarser power-of-two scales, which can make model quality more difficult to preserve.

\paragraph{Q4\_K.}
Q4\_K is an asymmetric weight-only integer format used by \texttt{llama.cpp}~\citep{gerganov2023llamacpp}. It arranges 256 weights into a super-block containing eight sub-blocks of 32 values. Within each sub-block, unsigned 4-bit integer codes represent the weights using a local scale and offset.

To reduce metadata overhead, Q4\_K applies double quantization to the sub-block scales $\bm{s}_6$ and offsets $\bm{m}_6$. These values are quantized to 6-bit integers and rescaled by the super-block FP16 parameters $s_{\mathrm{FP16}}$ and $m_{\mathrm{FP16}}$:
\begin{equation}
  \mathcal{Q}_{\mathrm{Q4\_K}}(\bm{W})
  \rightarrow
  \left(
    \bm{C}_{\mathrm{INT4}},
    \bm{s}_{6},
    \bm{m}_{6},
    s_{\mathrm{FP16}},
    m_{\mathrm{FP16}}
  \right).
  \label{eq:q4k}
\end{equation}
For weight \(i\) in sub-block \(b\), the dequantized value is:
\begin{equation}
  \hat{W}_{i}
  =
  \left(s_{\mathrm{FP16}} \cdot s_{6,b}\right) C_i
  -
  \left(m_{\mathrm{FP16}} \cdot m_{6,b}\right).
  \label{eq:q4k-dequant}
\end{equation}

\paragraph{Engine-specific scale sharing.}
Production inference engines such as vLLM and SGLang~\citep{vllm,sglang} fuse related projections into a single weight matrix, including the query, key, and value (QKV) projections in attention and the gate and up projections in the MLP. In NVFP4, each fused matrix uses a single tensor-level weight scale and a common input-activation scale. To reproduce this inference-time behavior, QATFactory ties the corresponding scales during training: it computes the shared weight scale from the concatenated projection weights and tracks one activation scale for their common input. Such scale tying is unnecessary for MXFP4 and Q4\_K because neither format uses tensor-level scales; their scales are defined at the block or super-block level.

\subsection{Quantization-aware distillation}
\label{sec:qad}

To recover accuracy lost to low-bit quantization, QATFactory uses self-distillation~\citep{hinton2015distilling,kim2019qkd,xin2026quantization}: the original high-precision pretrained checkpoint serves as a frozen teacher to supervise a low-precision student initialized from the same weights. Given an input sequence of length $L$, the framework minimizes the token-averaged forward Kullback--Leibler (KL) divergence between the teacher distribution $p_{T,t}$ and the student distribution $p_{S,t}$ over their shared vocabulary $\mathcal{V}$:
\begin{equation}
  \mathcal{L}_{\mathrm{QAD}}
  =
  \frac{1}{L}
  \sum_{t=1}^{L}
  \sum_{v \in \mathcal{V}}
  p_{T,t}(v)
  \log
  \frac{p_{T,t}(v)}{p_{S,t}(v)}.
  \label{eq:qad-objective}
\end{equation}

\paragraph{Quantization scope.}
The teacher remains frozen and runs in its original precision. In the student, QATFactory quantizes the linear projections to the target low-bit format and optimizes only their continuous latent weights. All other modules, including embeddings, normalization layers, routers, and the language-modeling head, remain frozen in their original precision to keep the student's behavior aligned with the teacher.

\paragraph{Full-parameter vs.\ LoRA-QAD.}
QATFactory supports two adaptation modes: full-parameter QAD and parameter-efficient QAD with low-rank adapters (LoRA). Full-parameter QAD maintains two separate copies of the model weights: frozen teacher weights and trainable student latent weights. It also maintains optimizer states for all student latent weights. This mode typically provides the highest quality, but has substantially higher memory requirements.

LoRA-QAD instead shares a single copy of the frozen base weights between the teacher and student and trains only the student's LoRA parameters. During the student forward pass, the LoRA parameters are merged with the base weights before quantization:
\begin{equation}
  \hat{\bm{W}}
  =
  \mathcal{Q}^{-1}
  \left(
    \mathcal{Q}
    \left(
      \bm{W}+\bm{B}\bm{A}
    \right)
  \right).
  \label{eq:qad-lora}
\end{equation}
Sharing the base weights and maintaining optimizer states only for the LoRA parameters significantly reduces memory requirements.

\paragraph{Mixture-of-experts adaptation.}
QATFactory supports both full-parameter and LoRA-QAD for mixture-of-experts (MoE) models~\citep{shazeer2017outrageously}. Full-parameter QAD directly trains the latent weights of every expert. LoRA-QAD supports two adapter layouts. In the \emph{unshared} layout, each expert \(e\) has its own low-rank factors \(\bm{A}_e\) and \(\bm{B}_e\). In the \emph{shared} (hybrid) layout, one factor is shared across experts while the other remains expert-specific: \(\bm{A}\) is shared for the gate and up projections, and \(\bm{B}\) is shared for the down projection.

To control peak memory during LoRA-QAD, QATFactory merges and quantizes one expert weight \((\bm{W}_e + \bm{B}_e\bm{A}_e)\) at a time during the forward pass. The temporary merged weight is released before the next expert is processed, avoiding the need to materialize high-precision merged weights for all experts simultaneously.

\subsection{Quantization-aware reinforcement learning}
\label{sec:qarl}

Quantization-aware reinforcement learning (QARL) extends deployment-aligned training to reinforcement learning with verifiable rewards (RLVR)~\citep{gu2026qarl}. QATFactory supports training algorithms such as GRPO~\citep{shao2024deepseekmath} and DAPO~\citep{yu2026dapo}.

\paragraph{Decoupled rollout and training servers.}
QATFactory uses separate servers for rollout generation and policy training. The rollout server loads the policy in the target low-bit format and uses an inference engine such as vLLM to generate responses with native low-precision kernels. Verifiers score these responses and send the resulting trajectories and rewards to the training server. The training server maintains high-precision latent weights and optimizer states in PyTorch. During each policy forward pass, it simulates the target quantization while performing the surrounding computation in BF16. Because rollout generation dominates step latency in the evaluated RLVR workload, native low-precision inference reduces generation time and improves end-to-end training throughput (\S\ref{sec:finding-qarl}).

\paragraph{Low-precision weight synchronization.}
After each training step, the training server quantizes the updated latent weights and packs the low-bit values and scale metadata in the inference engine's native layout. It then sends this representation to the rollout server. Sending inference-ready weights reduces the synchronization payload and avoids quantization or recalibration on the rollout server.



\subsection{Engine-native checkpoint export}
\label{sec:export}

QAD and QARL share an export pipeline that produces low-precision checkpoints in the target inference engine's native format.

\paragraph{Preparing weights for export.}
For LoRA-QAD, the exporter first merges the trained adapters into the frozen base weights and then quantizes the merged weights:
\begin{equation}
  \bm{W}_{\mathrm{export}}
  =
  \mathcal{Q}
  \left(
    \bm{W}+\bm{B}\bm{A}
  \right),
  \label{eq:export-lora}
\end{equation}
This produces a quantized checkpoint from the same merged weights used during training. For full-parameter QAD and QARL, the exporter instead quantizes the final high-precision latent weights directly.

\paragraph{Packing the checkpoint.}
The exporter packs the low-bit values, scale factors, and format-specific metadata in the target engine's native layout. It uses the same quantization configuration as training. When a format requires tracked state, such as NVFP4 input-activation scales, the exporter saves the values collected during training instead of running a separate calibration pass. The resulting checkpoint can be loaded directly by engines such as vLLM, SGLang, or \texttt{llama.cpp}, without additional post-training quantization or calibration.

\section{Experiments and Findings}
\label{sec:findings}

Using \projectname, we conduct a series of quantization-aware distillation (QAD) and quantization-aware reinforcement learning (QARL) experiments across diverse model families, scales, and quantization formats. We evaluate the trained checkpoints directly in production inference engines, compare their benchmark accuracy with post-training quantization (PTQ) baselines, and measure distribution fidelity relative to the unquantized reference. Below, we describe the experimental setup and present our five empirical findings.

\begin{figure*}[!t]
    \centering
    \includegraphics[width=\textwidth]{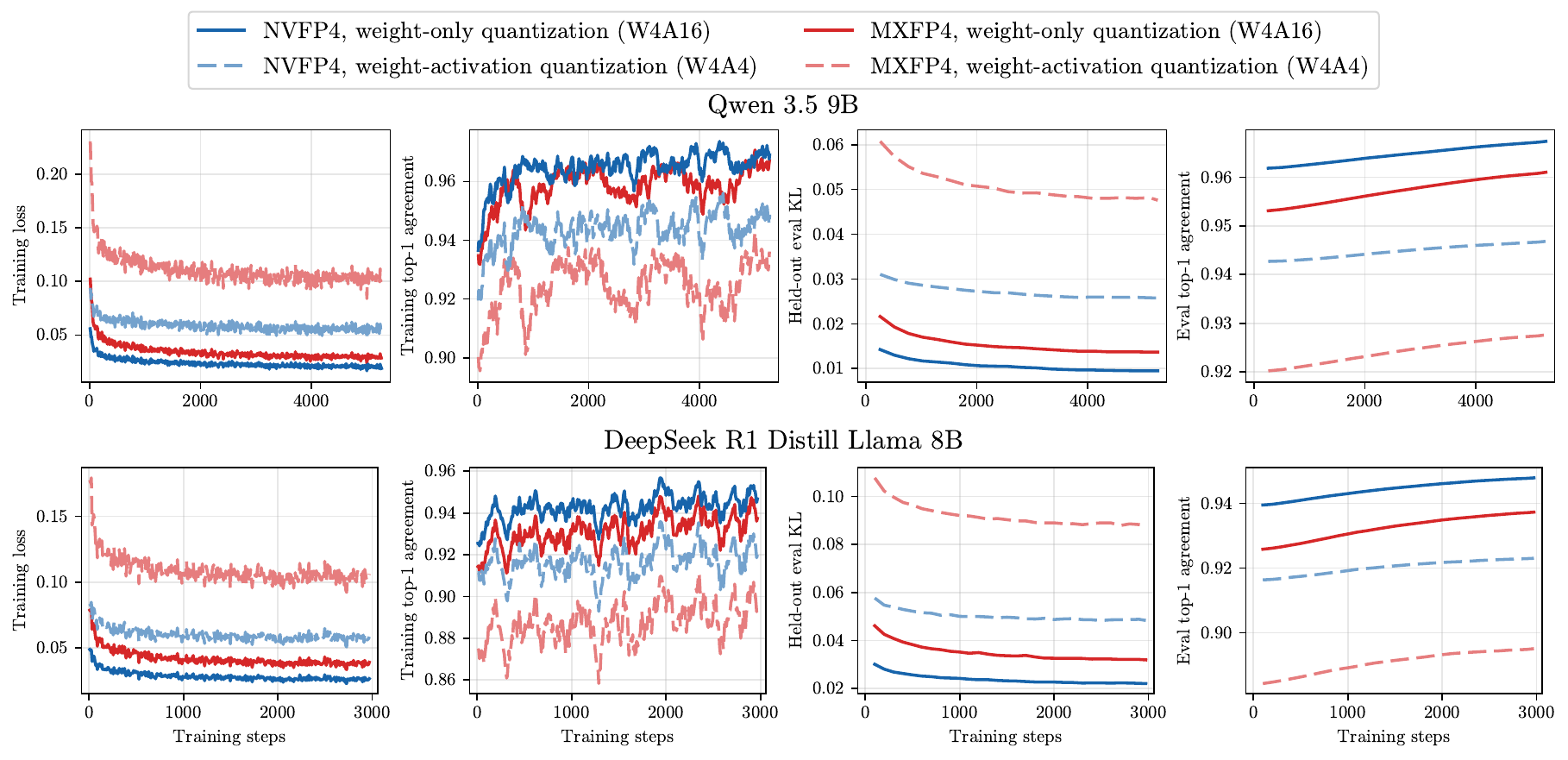}
    \caption{Training loss, training top-1 agreement, evaluation KL on held-out data, and evaluation top-1 agreement for Qwen3.5 9B and DeepSeek R1 Distill Llama 8B. We compare weight-only (W4A16) and weight--activation (W4A4) training under NVFP4 and MXFP4.}
    \label{fig:training-nvfp4-mxfp4}
\end{figure*}

\subsection{Experimental Setup}
\label{sec:setup}

\begin{table*}[t]
    \centering
    \small
    \setlength{\tabcolsep}{3pt}
    \begin{adjustbox}{max width=\textwidth}
    \begin{tabular}{l rrrrrrrr @{\hspace{0.6em}}r @{\hspace{0.9em}} rrr @{\hspace{0.6em}}r}
      \toprule
      & \multicolumn{9}{c}{\textbf{Benchmark accuracy} ($\uparrow$)}
      & \multicolumn{4}{c}{\textbf{KL divergence} ($\downarrow$)} \\
      \cmidrule(lr){2-10} \cmidrule(lr){11-14}
      \textbf{Checkpoint} & GPQA-D & MMLU-P & MMMLU & WinoG. & HellaS.
      & AIME25 & LCB & BCB & \textbf{Avg}
      & UltraChat & Pile & WikiText & \textbf{Avg} \\
      \midrule
      \multicolumn{14}{l}{\textbf{Qwen 3.5 9B}} \\
      \quad BF16 (reference) & 84.3 & 72.8 & 87.4 & 90.3 & 74.0 & 50.7 & 82.3 & 31.7 & 71.7
        & --- & --- & --- & --- \\
      \addlinespace
      \multicolumn{14}{l}{\quad \textit{NVFP4 (W4A4)}} \\
      \quad\quad RTN & 77.8 & 72.1 & \textbf{84.4} & \textbf{89.3} & 70.0 & 40.7 & 63.7 & 25.0 & 65.4
        & 0.0948 & 0.0566 & 0.0820 & 0.0778 \\
      \quad\quad GPTQ & 71.2 & 71.8 & 80.6 & 88.7 & 70.7 & 34.0 & 66.0 & 23.3 & 63.3
        & 0.2121 & 0.0894 & 0.1181 & 0.1399 \\
      \rowcolor{qadrow} \quad\quad QAD (trained with W4A16) & 77.8 & \textbf{73.5} & 82.7 & \textbf{89.3} & 72.0 & 43.3 & \textbf{83.0} & \textbf{29.7} & \textbf{68.9}
        & \textbf{0.0852} & \textbf{0.0484} & \textbf{0.0710} & \textbf{0.0682} \\
      \rowcolor{qadrow} \quad\quad QAD (trained with W4A4) & \textbf{78.3} & 71.4 & 83.7 & 89.0 & \textbf{75.0} & \textbf{47.3} & 78.3 & 28.0 & \textbf{68.9}
        & 0.1008 & 0.0507 & 0.0723 & 0.0746 \\
      \addlinespace
      \multicolumn{14}{l}{\quad \textit{MXFP4 (W4A4)}} \\
      \quad\quad RTN & 33.8 & 62.6 & 71.8 & 84.0 & 70.3 & 1.3 & 15.3 & 15.3 & 44.3
        & 0.2303 & 0.1331 & 0.1742 & 0.1792 \\
      \quad\quad GPTQ & 63.1 & 68.0 & \textbf{82.3} & 87.0 & 68.3 & 27.3 & 39.0 & 16.3 & 56.4
        & \textbf{0.1800} & 0.1234 & 0.1670 & 0.1568 \\
      \rowcolor{qadrow} \quad\quad QAD (trained with W4A16) & 66.7 & 68.7 & 79.2 & 86.7 & 71.3 & 28.7 & 47.7 & 13.3 & 57.8
        & 0.1942 & 0.1109 & 0.1478 & 0.1510 \\
      \rowcolor{qadrow} \quad\quad QAD (trained with W4A4) & \textbf{74.8} & \textbf{71.8} & 81.3 & \textbf{89.7} & \textbf{74.0} & \textbf{44.0} & \textbf{72.3} & \textbf{20.0} & \textbf{66.0}
        & 0.2031 & \textbf{0.0980} & \textbf{0.1303} & \textbf{0.1438} \\
      \midrule
      \multicolumn{14}{l}{\textbf{DeepSeek R1 Distill Llama 8B}} \\
      \quad BF16 (reference) & 50.3 & 54.3 & 54.6 & 68.2 & 52.8 & 35.0 & 64.2 & 15.8 & 49.4
        & --- & --- & --- & --- \\
      \addlinespace
      \multicolumn{14}{l}{\quad \textit{NVFP4 (W4A4)}} \\
      \quad\quad RTN & 45.3 & 48.9 & 52.3 & 67.7 & 53.5 & 30.8 & 57.9 & 13.2 & 46.2
        & 0.1235 & 0.1596 & 0.1866 & 0.1566 \\
      \quad\quad GPTQ & 45.7 & 49.8 & 50.6 & 68.6 & 52.8 & 32.0 & 57.4 & 14.3 & 46.4
        & 0.1086 & 0.1335 & 0.1580 & 0.1334 \\
      \rowcolor{qadrow} \quad\quad QAD (trained with W4A16) & \textbf{46.1} & \textbf{50.4} & \textbf{54.0} & \textbf{68.9} & \textbf{55.3} & \textbf{32.7} & \textbf{58.5} & \textbf{15.3} & \textbf{47.7}
        & \textbf{0.0855} & \textbf{0.1174} & \textbf{0.1311} & \textbf{0.1113} \\
      \rowcolor{qadrow} \quad\quad QAD (trained with W4A4) & 43.9 & 48.3 & 53.4 & 68.2 & 54.5 & 32.0 & 57.8 & 15.0 & 46.6
        & 0.0891 & 0.1197 & 0.1356 & 0.1148 \\
      \addlinespace
      \multicolumn{14}{l}{\quad \textit{MXFP4 (W4A4)}} \\
      \quad\quad RTN & 39.8 & 42.2 & 46.1 & 62.9 & \textbf{54.1} & 26.0 & 50.2 & 12.8 & 41.8
        & 0.2548 & 0.3535 & 0.3880 & 0.3321 \\
      \quad\quad GPTQ & 40.1 & 45.2 & 46.1 & 62.9 & 52.4 & 26.3 & 49.7 & 12.2 & 41.9
        & 0.2215 & 0.2910 & 0.3249 & 0.2791 \\
      \rowcolor{qadrow} \quad\quad QAD (trained with W4A16) & 42.5 & 45.9 & 48.1 & 65.3 & 52.0 & \textbf{29.2} & \textbf{53.3} & \textbf{13.6} & 43.7
        & 0.1787 & 0.2374 & 0.2754 & 0.2305 \\
      \rowcolor{qadrow} \quad\quad QAD (trained with W4A4) & \textbf{43.6} & \textbf{47.4} & \textbf{48.8} & \textbf{67.0} & 54.0 & 25.8 & 52.2 & 13.0 & \textbf{44.0}
        & \textbf{0.1601} & \textbf{0.2143} & \textbf{0.2525} & \textbf{0.2090} \\
      \bottomrule
    \end{tabular}
    \end{adjustbox}
    \caption{Benchmark accuracy and KL divergence to the BF16 reference for Qwen3.5 9B and DeepSeek R1 Distill Llama 8B. All quantized checkpoints are deployed at W4A4. The QAD rows differ only in whether activations are quantized during training. Higher accuracy and lower KL divergence are better.}
    \label{tab:main-qad-results}
\end{table*}

\paragraph{Quantization-aware distillation (QAD).}
We construct distillation data from Open Perfect Blend~\citep{perfectblend} by pairing prompts with responses generated by the original unquantized model. We train at an 8K context length using the forward KL divergence between the full-precision and quantized models. We train using the AdamW optimizer~\citep{loshchilov2017decoupled} with a decaying learning rate of \(1\times10^{-6}\) with warmup and a batch size of 16. Most QAD runs are performed on a single node of 8 NVIDIA H100 GPUs, whereas Qwen3-30B-A3B and MiniMax M2.7 are trained on 16 and 32 H100 GPUs, respectively.

\paragraph{Quantization-aware reinforcement learning (QARL).}
For reinforcement learning, we train Qwen3-8B-Base~\citep{yang2025qwen3} on DeepMath-103K~\citep{he2026deepmath} for 1,024 steps using GRPO~\citep{shao2024deepseekmath} with the DAPO loss~\citep{yu2026dapo}. In NVFP4 QARL, vLLM~\citep{vllm} executes rollouts directly with native NVFP4 weights, sampling 16 rollouts per prompt. Training workers compute policy updates under simulated quantization and synchronize updated quantized weights to the rollout engine after each step. We use a constant learning rate of \(1\times10^{-6}\) following an initial warmup. Because native NVFP4 rollout execution requires Blackwell-generation Tensor Cores~\citep{nvidia_cuda_12_8}, QARL experiments are conducted on 4 NVIDIA B200 GPUs.

\paragraph{Models and quantization formats.}
We train and evaluate dense models (Qwen3.5 9B~\citep{qwen3.5}, DeepSeek R1 Distill Llama 8B~\citep{guo2025deepseek}, and Muse-Glimmer 30B~\citep{meta2026museglimmer}) and mixture-of-experts (MoE) models (Qwen3-30B-A3B~\citep{yang2025qwen3} and MiniMax M2.7~\citep{chen2026minimax}), spanning 8B to 230B total parameters. We target three low-precision formats: NVFP4, MXFP4, and llama.cpp's Q4\_K. For W4A4 deployment formats, including NVFP4 and MXFP4, we compare W4A16 training (weight-only quantization) with W4A4 training (weight-activation quantization).

\paragraph{Model Quality Evaluations.}
During training, we record training loss, teacher--student top-1 token agreement, and the forward KL divergence on 128 held-out samples. After training, all checkpoints are exported to their target engine formats (vLLM and llama.cpp) without additional calibration. We report downstream benchmark accuracy on GPQA-Diamond~\citep{rein2024gpqa}, MMLU-Pro~\citep{wang2024mmlu}, MMMLU~\citep{mmmlu}, Winogrande~\citep{sakaguchi2021winogrande}, HellaSwag~\citep{zellers2019hellaswag}, AIME25~\citep{aime25}, LiveCodeBench~\citep{jain2025livecodebench}, and BigCodeBench~\citep{zhuo2025bigcodebench}. We measure distribution fidelity via KL divergence to the unquantized reference on UltraChat~\citep{ultrachat}, the Pile~\citep{gao2020pile}, and WikiText~\citep{wikitext}. We compare exported QAD checkpoints against round-to-nearest (RTN) and GPTQ~\citep{frantar2022gptq} post-training quantization baselines. For Q4\_K, we use \texttt{llama.cpp}'s default MSE-optimal PTQ algorithm~\citep{gerganov2023llamacpp} as the baseline. This algorithm searches for scales that minimize \(\|W - W_q\|_2^2\), where \(W\) and \(W_q\) are the original and dequantized weight matrices.

\newtcolorbox{findingbox}{
    colback=qadrow,
    colframe=blue!40!black,
    boxrule=0.6pt,
    arc=2pt,
    left=8pt,
    right=8pt,
    top=6pt,
    bottom=6pt,
    before skip=10pt,
    after skip=6pt,
    fontupper=\bfseries
}

\subsection{Generality Across Formats, Scales, and Architectures}
\label{sec:finding-generality}

\begin{findingbox}
Finding 1: \projectname converges stably across model families, parameter counts from 8B to 230B, and NVFP4, MXFP4, and Q4\_K. In vLLM and \texttt{llama.cpp}, the exported QAD checkpoints achieve higher average benchmark accuracy and lower mean KL divergence than the RTN and GPTQ post-training quantization baselines.
\end{findingbox}

\begin{figure*}[!t]
    \centering
    \includegraphics[width=\textwidth]{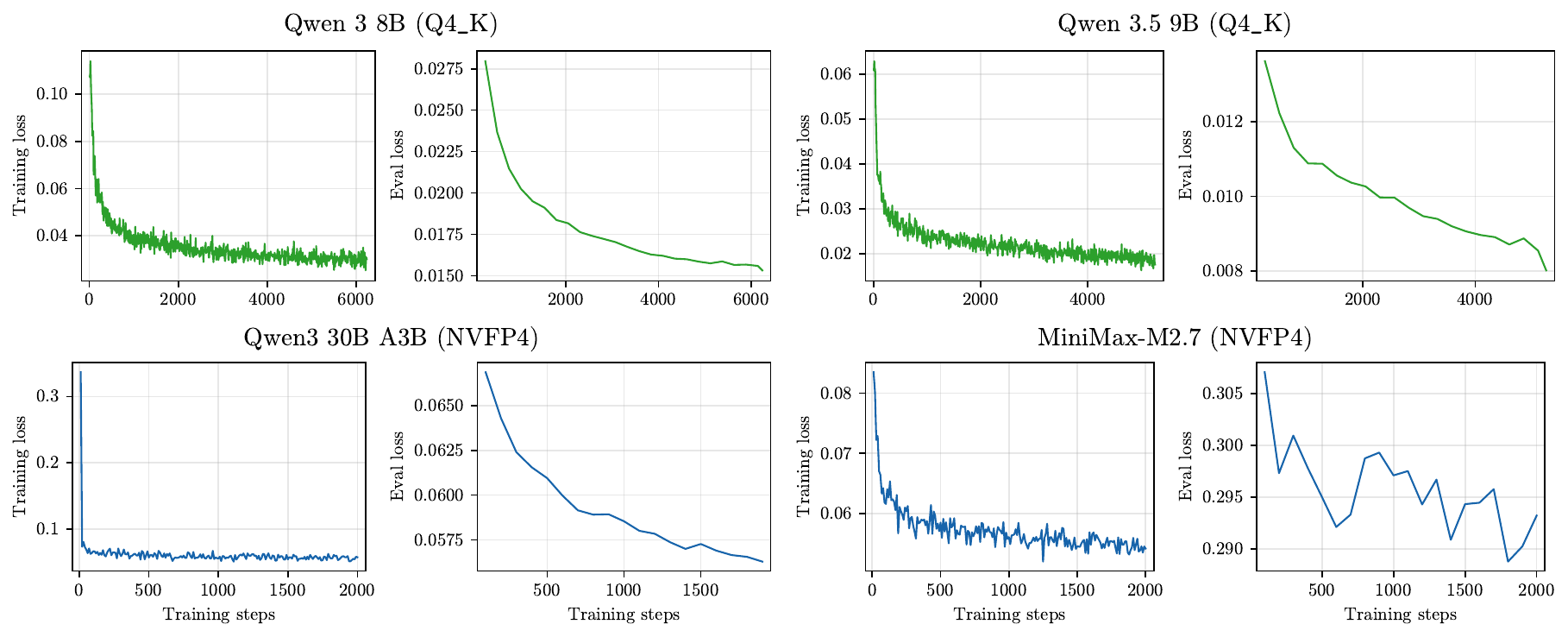}
    \caption{Training and held-out evaluation loss for Qwen3 8B and Qwen3.5 9B under Q4\_K quantization, and for Qwen3-30B-A3B and MiniMax M2.7 under NVFP4 quantization.}
    \label{fig:training-additional-models}
\end{figure*}

Figures~\ref{fig:training-nvfp4-mxfp4} and~\ref{fig:training-additional-models} illustrate training and evaluation curves across dense and MoE architectures under NVFP4, MXFP4, and Q4\_K. For Qwen3.5 9B and DeepSeek R1 Distill Llama 8B, training loss and evaluation KL divergence decrease steadily while top-1 agreement rises. Q4\_K training converges smoothly for both Qwen3 8B and Qwen3.5 9B. On larger MoE models, NVFP4 QAD trains stably on both Qwen3-30B-A3B and the 230B-scale MiniMax M2.7. The same framework and training recipe converge across these numerical formats and model architectures.

As shown in Table~\ref{tab:main-qad-results}, QAD consistently outperforms the PTQ baselines across formats and model families, with higher average accuracy and lower KL divergence. The differences are especially large on coding benchmarks. For Qwen3.5 9B under NVFP4, QAD achieves 83.0\% on LiveCodeBench (LCB), compared with 63.7\% for RTN and 66.0\% for GPTQ, and 29.7\% on BigCodeBench (BCB), compared with 25.0\% for RTN and 23.3\% for GPTQ. The QAD scores on LCB and BCB are close to the corresponding BF16 scores of 82.3\% and 31.7\%. Under MXFP4, QAD achieves 72.3\% on LCB and an average benchmark accuracy of 66.0\%, whereas the best PTQ baseline achieves 39.0\% and 56.4\%, respectively. On DeepSeek R1 Distill Llama 8B, QAD also has the highest average accuracy under both NVFP4 (47.7\% vs.\ 46.4\% for the best PTQ baseline) and MXFP4 (44.0\% vs.\ 41.9\%).

Table~\ref{tab:moe-results} shows the same pattern for large MoEs and edge formats. Under NVFP4, QAD improves average accuracy over PTQ by 1.5 percentage points on Qwen3-30B-A3B and 2.5 points on MiniMax M2.7. It also lowers mean KL divergence on UltraChat, the Pile, and WikiText. Under \texttt{llama.cpp}'s Q4\_K format, QAD outperforms MSE-optimal PTQ on both evaluated models. For Qwen3.5 9B, QAD achieves 49.33\% on AIME25, compared with 38.00\% for PTQ, and its mean KL divergence is 25--30\% lower.

\begin{table*}[!t]
    \centering
    \small
    \setlength{\tabcolsep}{4pt}
    \begin{adjustbox}{max width=\textwidth}
    \begin{tabular}{l rrrrr @{\hspace{0.9em}} rrr @{\hspace{0.6em}}r}
      \toprule
      & \multicolumn{5}{c}{\textbf{Benchmark accuracy} ($\uparrow$)}
      & \multicolumn{4}{c}{\textbf{KL divergence} ($\downarrow$)} \\
      \cmidrule(lr){2-6} \cmidrule(lr){7-10}
      \textbf{Checkpoint} & GPQA-D & AIME25 & BCB & LCB & \textbf{Avg}
      & UltraChat & Pile & WikiText & \textbf{Avg} \\
      \midrule
      \multicolumn{10}{l}{\textbf{Qwen3-30B-A3B (NVFP4)}} \\
      \quad BF16 (reference) & 62.12 & 73.33 & 39.00 & 93.00
        & 66.86 & --- & --- & --- & --- \\
      \quad RTN & 59.09 & 69.33 & \textbf{38.33} & 90.33
        & 64.27 & 0.03813 & 0.02660 & 0.03624 & 0.03417 \\
      \rowcolor{qadrow}
      \quad QAD & \textbf{61.11} & \textbf{72.67}
        & \textbf{38.33} & \textbf{91.00}
        & \textbf{65.78}
        & \textbf{0.03381} & \textbf{0.02489} & \textbf{0.03545}
        & \textbf{0.03242} \\
      \midrule
      \multicolumn{10}{l}{\textbf{MiniMax M2.7 (NVFP4)}} \\
      \quad FP8 (reference) & 78.8 & 78.0 & 22.0 & 88.7
        & 66.9 & --- & --- & --- & --- \\
      \quad RTN & 80.3 & 78.0 & 22.0 & 86.0
        & 66.6 & 0.0517 & 0.0474 & 0.1033 & 0.0779 \\
      \rowcolor{qadrow}
      \quad QAD & \textbf{82.8} & \textbf{82.7}
        & \textbf{23.3} & \textbf{87.7}
        & \textbf{69.1}
        & \textbf{0.0454} & \textbf{0.0418} & \textbf{0.0930}
        & \textbf{0.0697} \\
      \midrule
      \multicolumn{10}{l}{\textbf{Qwen3-8B (Q4\_K)}} \\
      \quad BF16 (reference) & 60.10 & 66.67 & --- & --- & 63.39 & --- & --- & --- & --- \\
      \quad MSE-optimal PTQ & 56.06 & 62.67 & --- & --- & 59.37 & --- & --- & 0.0628 & --- \\
      \rowcolor{qadrow}
      \quad QAD & \textbf{59.09} & \textbf{63.33} & --- & --- & \textbf{61.21} & --- & --- & \textbf{0.0436} & --- \\
      \midrule
      \multicolumn{10}{l}{\textbf{Qwen3.5-9B (Q4\_K)}} \\
      \quad BF16 (reference) & 85.86 & 54.00 & --- & --- & 69.93 & --- & --- & --- & --- \\
      \quad MSE-optimal PTQ & 77.27 & 38.00 & --- & --- & 57.64 & --- & --- & 0.0835 & --- \\
      \rowcolor{qadrow}
      \quad QAD & \textbf{79.80} & \textbf{49.33} & --- & --- & \textbf{64.57} & --- & --- & \textbf{0.0623} & --- \\
      \bottomrule
    \end{tabular}
    \end{adjustbox}
    \caption{Benchmark accuracy and KL divergence for MoE architectures under NVFP4 and dense architectures under \texttt{llama.cpp}'s Q4\_K format. Dashes indicate benchmarks not evaluated for that configuration. Higher accuracy and lower KL divergence are better.}
    \label{tab:moe-results}
\end{table*}

\subsection{Choosing Training-Time Activation Precision for W4A4 Deployment}
\label{sec:finding-activation}

\begin{findingbox}
Finding 2: For W4A4 deployment, quantizing activations during training is not always optimal. Although NVFP4 inference uses W4A4, W4A16 training produces higher-quality quantized checkpoints than W4A4 training; by contrast, MXFP4 benefits substantially from W4A4 training.
\end{findingbox}

Table~\ref{tab:main-qad-results} shows that the preferred training precision depends on the format. For NVFP4, W4A16 matches or outperforms W4A4 on both models. For MXFP4, W4A4 substantially improves Qwen3.5 9B, raising average accuracy from 57.8\% to 66.0\% and reducing average KL divergence from 0.1510 to 0.1438.

This format-dependent behavior reflects a trade-off between exposing the model to deployment-time activation quantization during training and preserving a cleaner gradient signal. Consider a linear layer with input activation $\bm{A}$, quantized weight $\hat{\bm{W}}$, and output $\bm{O}$. During W4A4 training, the activation quantize--dequantize operation produces
\begin{equation}
  \hat{\bm{A}}
  =
  \mathcal{Q}(\bm{A})
  =
  \bm{A} + \bm{\varepsilon}_{A},
  \qquad
  \bm{O}
  =
  \hat{\bm{W}}\hat{\bm{A}},
  \label{eq:activation-quantization}
\end{equation}
where $\bm{\varepsilon}_{A}$ denotes the activation quantization error. During W4A16 training, activations remain in high precision, so $\hat{\bm{A}}=\bm{A}$ and $\bm{\varepsilon}_{A}=\bm{0}$.

As with weight quantization in Equation~\ref{eq:ste}, the backward pass uses an STE for the activation quantizer:
\begin{equation}
  \frac{\partial \hat{\bm{A}}}{\partial \bm{A}}
  \approx
  \bm{I},
  \qquad
  \frac{\partial \mathcal{L}}{\partial \bm{A}}
  \approx
  \hat{\bm{W}}^{\mathsf{T}}
  \frac{\partial \mathcal{L}}{\partial \bm{O}}.
  \label{eq:activation-ste}
\end{equation}
The STE passes gradients through the discrete quantizer by treating it as the identity function, but it does not remove the error introduced during the forward pass. Under W4A4 training, the approximate weight gradient is
\begin{equation}
  \frac{\partial \mathcal{L}}{\partial \bm{W}}
  \approx
  \frac{\partial \mathcal{L}}{\partial \bm{O}}
  \hat{\bm{A}}^{\mathsf{T}}
  =
  \frac{\partial \mathcal{L}}{\partial \bm{O}}
  \bm{A}^{\mathsf{T}}
  +
  \frac{\partial \mathcal{L}}{\partial \bm{O}}
  \bm{\varepsilon}_{A}^{\mathsf{T}}.
  \label{eq:activation-gradient-error}
\end{equation}
For a fixed output gradient, the second term captures the direct perturbation to the weight update. Activation quantization also changes the forward output and therefore $\partial\mathcal{L}/\partial\bm{O}$. Thus, activation quantization affects the gradient both directly through $\bm{\varepsilon}_A$ and indirectly through the perturbed output gradient. W4A4 training improves alignment with deployment-time activation errors but makes teacher matching more difficult by optimizing from perturbed activations. The higher training loss in Figure~\ref{fig:training-nvfp4-mxfp4} is consistent with this trade-off.

One possible explanation is the difference in how the two formats scale activation blocks. NVFP4 uses 16-element blocks with E4M3 block scales and an FP32 tensor scale, which may keep activation quantization error relatively small. In this case, quantizing activations during training may add noise without providing much benefit, allowing W4A16-trained weights to remain effective after W4A4 export. MXFP4 uses larger, 32-element blocks with power-of-two E8M0 scales, so the gap between high-precision training activations and quantized deployment activations may be larger. W4A4 training may help the model adapt to this gap, even if it makes optimization noisier. Although this explanation remains tentative, the results suggest choosing training-time activation precision separately for each format and evaluating the exported checkpoints.

\begin{table}[t]
  \centering
  \small
  \begin{tabular}{lrrr}
    \toprule
    \textbf{Category} & \textbf{Full-param QAD} & \textbf{LoRA $r{=}16$} & \textbf{$\Delta$} \\
    \midrule
    Teacher weights (BF16)                     & 16.68 & 0.00  & $-$16.68 \\
    Student weights, frozen (BF16)             & 3.79  & 16.68 & $+$12.89 \\
    Student weights, trainable (FP32)          & 25.77 & 0.16  & $-$25.61 \\
    Gradients (FP32)                           & 25.77 & 0.16  & $-$25.61 \\
    Optimizer states (FP32)                    & 51.55 & 0.32  & $-$51.22 \\
    \textbf{Persistent subtotal}               & \textbf{123.56} & \textbf{17.32} & \textbf{$-$106.24} \\
    \midrule
    Activations                                & 32.97 & 35.37 & $+$2.40 \\
    Temporary                                  & 10.00 & 4.50  & $-$5.50 \\
    \midrule
    \textbf{GPU memory in use}                 & \textbf{166.53} & \textbf{57.19} & \textbf{$-$109.34} \\
    \bottomrule
  \end{tabular}
  \caption{Training memory breakdown (GiB) measured for Qwen3.5 9B with an 8K sequence length, comparing full-parameter QAD with LoRA QAD ($r=16$) under the same hardware configuration. $\Delta$ is the LoRA value minus the full-parameter QAD value. LoRA removes the separate teacher copy and confines trainable parameters, gradients, and optimizer states to the adapters. In this configuration, LoRA reduces GPU memory usage by roughly $2.9\times$.}
  \label{tab:memory}
\end{table}

\subsection{Parameter-Efficient QAT via LoRA: Memory Footprint and Scaling Limits}
\label{sec:finding-lora}

\begin{findingbox}
Finding 3: LoRA significantly reduces training memory requirements: for Qwen3.5 9B at an 8K sequence length, rank-16 LoRA reduces GPU memory usage by \(2.9\times\). However, LoRA underperforms full-parameter QAD, and increasing the adapter rank does not close the gap.
\end{findingbox}

Table~\ref{tab:memory} reports the memory footprint of LoRA-QAD (\(r=16\)) versus full-parameter QAD when training Qwen3.5 9B with an 8K sequence length. In this configuration, LoRA reduces GPU memory usage by \(2.91\times\), from around \(167\)\,GiB to \(57\)\,GiB. This reduction comes primarily from model weights and optimizer states: freezing the base model eliminates the need for a separate teacher copy and restricts trainable weights, gradients, and FP32 AdamW optimizer states to low-rank adapters (\(17.32\)\,GiB vs.\ \(123.56\)\,GiB persistent subtotal). Activation memory remains comparable (\(\sim 35.4\)\,GiB vs.\ \(\sim 33.0\)\,GiB) because both approaches execute the same forward--backward passes.

\begin{figure*}[t]
    \centering
    \includegraphics[width=0.8\textwidth]{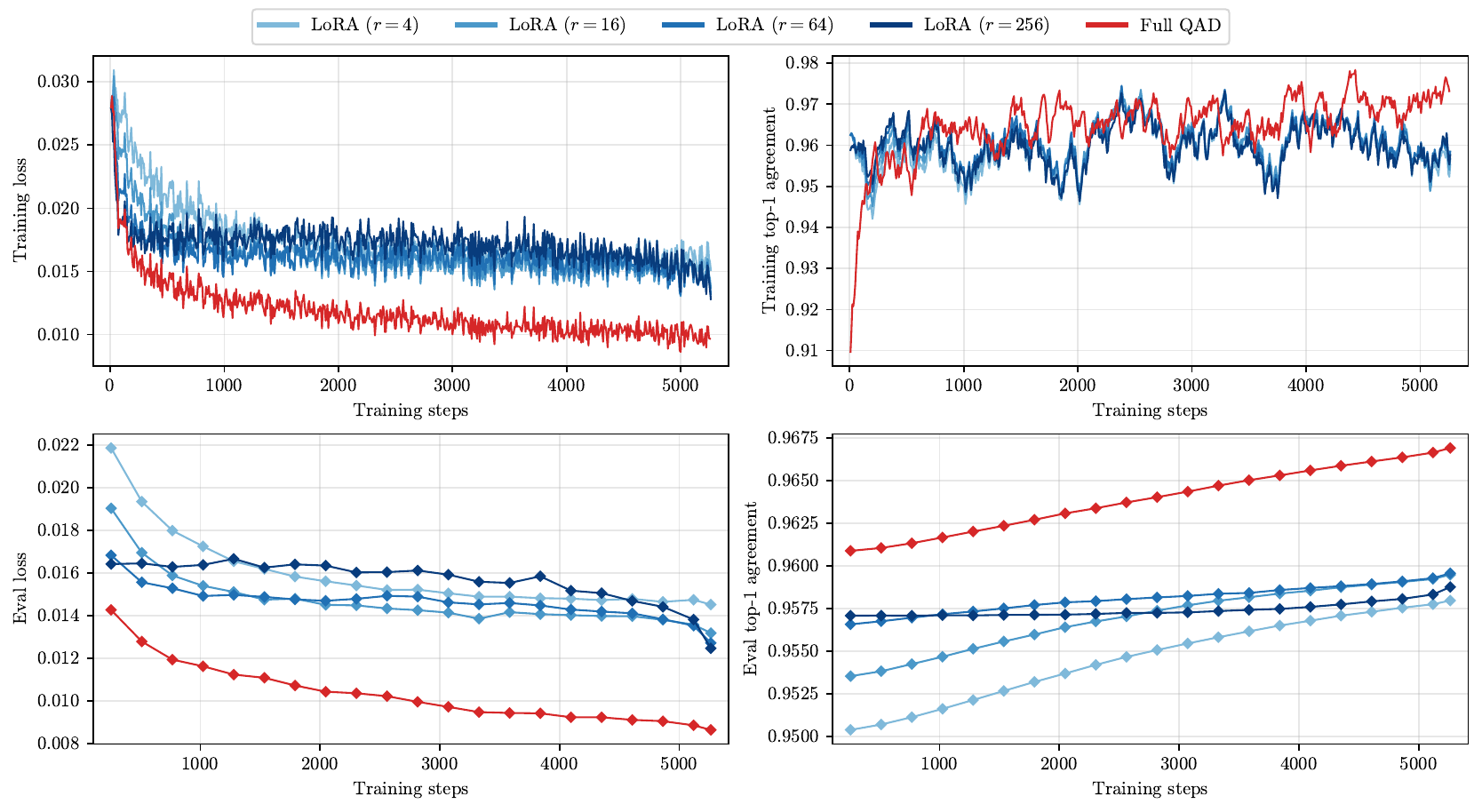}
    \caption{Training loss, training top-1 agreement, held-out evaluation loss, and held-out top-1 agreement for LoRA QAD at ranks 4, 16, 64, and 256, compared with full-parameter QAD on Qwen3.5 9B.}
    \label{fig:lora-training}
\end{figure*}

\begin{table*}[!t]
    \centering
    \small
    \setlength{\tabcolsep}{3pt}
    \begin{adjustbox}{max width=\textwidth}
    \begin{tabular}{l rrrrrrrr @{\hspace{0.6em}}r @{\hspace{0.9em}} rrr @{\hspace{0.6em}}r}
      \toprule
      & \multicolumn{9}{c}{\textbf{Benchmark accuracy} ($\uparrow$)}
      & \multicolumn{4}{c}{\textbf{KL divergence} ($\downarrow$)} \\
      \cmidrule(lr){2-10} \cmidrule(lr){11-14}
      \textbf{Checkpoint} & GPQA-D & MMLU-P & MMMLU & WinoG. & HellaS.
      & AIME25 & LCB & BCB & \textbf{Avg}
      & UltraChat & Pile & WikiText & \textbf{Avg} \\
      \midrule
      BF16 (reference) & 84.3 & 72.8 & 87.4 & 90.3 & 74.0 & 50.7 & 82.3 & 31.7 & 71.7
        & --- & --- & --- & --- \\
      \addlinespace
      \multicolumn{14}{l}{\textit{NVFP4}} \\
      \quad Full QAD & 77.8 & 73.5 & 82.7 & 89.3 & 72.0 & 43.3 & \textbf{83.0} & 29.7 & \textbf{68.9}
        & 0.0852 & \textbf{0.0484} & \textbf{0.0710} & \textbf{0.0682} \\
      \quad QAD $+$ LoRA ($r{=}4$) & \textbf{78.8} & 74.2 & 84.4 & 90.0 & \textbf{75.0} & 40.0 & 70.0 & \textbf{31.7} & 68.0
        & \textbf{0.0831} & 0.0538 & 0.0797 & 0.0722 \\
      \quad QAD $+$ LoRA ($r{=}16$) & 77.8 & 72.8 & 83.3 & \textbf{90.7} & 70.3 & 45.3 & 73.0 & 29.3 & 67.8
        & 0.0929 & 0.0520 & 0.0756 & 0.0735 \\
      \quad QAD $+$ LoRA ($r{=}64$) & 76.8 & 73.5 & 82.0 & 88.7 & 73.0 & \textbf{46.7} & 75.7 & 27.0 & 67.9
        & 0.0885 & 0.0519 & 0.0738 & 0.0714 \\
      \quad QAD $+$ LoRA ($r{=}256$) & 73.2 & \textbf{74.5} & \textbf{84.7} & 89.3 & \textbf{75.0} & \textbf{46.7} & 73.0 & 24.7 & 67.6
        & 0.1056 & 0.0529 & 0.0747 & 0.0777 \\
      \bottomrule
    \end{tabular}
    \end{adjustbox}
    \caption{Benchmark accuracy and KL divergence for LoRA QAD on Qwen3.5 9B at increasing adapter rank, compared with full-parameter QAD. Higher accuracy and lower KL divergence are better.}
    \label{tab:lora-results}
\end{table*}

\begin{figure*}[!t]
  \centering
  \includegraphics[width=0.96\textwidth]{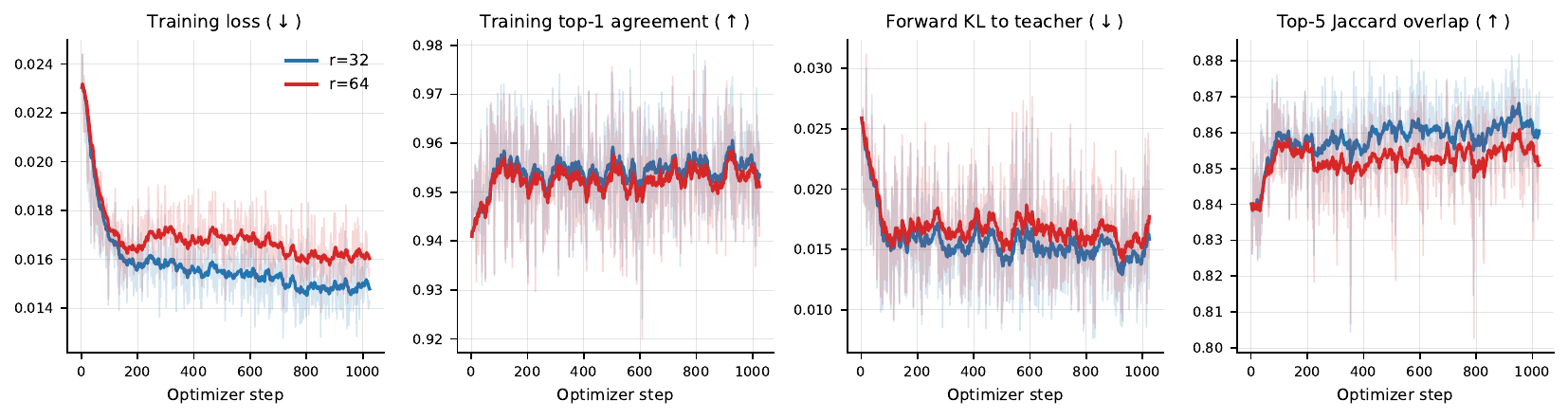}
  \caption{LoRA QAD of Muse-Glimmer 30B at ranks 32 and 64: training loss,
  training top-1 agreement, forward KL divergence to the teacher, and top-5
  Jaccard overlap. Faint lines show per-step values, and solid lines show their
  exponential moving averages.}
  \label{fig:lora_muse}
\end{figure*}

We evaluate LoRA-QAD at multiple adapter ranks on two models: Qwen3.5 9B with \(r \in \{4,16,64,256\}\) and Muse-Glimmer 30B with \(r \in \{32,64\}\), both under NVFP4 quantization. Figures~\ref{fig:lora-training} and~\ref{fig:lora_muse} show their training and evaluation metrics. Tables~\ref{tab:lora-results} and~\ref{tab:lora_muse} report benchmark accuracy and KL divergence for the exported checkpoints after evaluation in vLLM.

For Qwen3.5 9B, Figure~\ref{fig:lora-training} shows that LoRA-QAD recovers model quality over the course of training: held-out evaluation loss decreases and top-1 agreement increases. However, model quality does not improve monotonically with adapter rank. Table~\ref{tab:lora-results} shows that rank 4 achieves the highest average benchmark accuracy among the LoRA configurations (68.0\%), rank 64 achieves the lowest average KL divergence (0.0714), and rank 256 yields a lower average accuracy of 67.6\%. On the aggregate metrics, all LoRA configurations remain behind full-parameter QAD, which achieves 68.9\% average accuracy and 0.0682 average KL divergence.

\begin{table}[!t]
  \centering
  \small
  \setlength{\tabcolsep}{3pt}
  \begin{adjustbox}{max width=\textwidth}
  \begin{tabular}{l rrrrrrrr @{\hspace{0.6em}}r @{\hspace{0.9em}} rrr @{\hspace{0.6em}}r}
    \toprule
    & \multicolumn{9}{c}{\textbf{Benchmark accuracy} ($\uparrow$)}
    & \multicolumn{4}{c}{\textbf{KL divergence} ($\downarrow$)} \\
    \cmidrule(lr){2-10} \cmidrule(lr){11-14}
    \textbf{Checkpoint} & GPQA-D & MMLU-P & MMMLU & WinoG. & HellaS.
    & AIME25 & LCB & BCB & \textbf{Avg}
    & UltraChat & Pile & WikiText & \textbf{Avg} \\
    \midrule
    BF16 (reference) & 85.4 & 79.9 & 90.5 & 92.3 & 68.0 & 86.0 & 85.3 & 41.0 & 78.6
      & --- & --- & --- & --- \\
    \addlinespace
    \multicolumn{14}{l}{\textit{NVFP4}} \\
    \quad RTN & 79.3 & 79.9 & 89.1 & 93.0 & 63.7 & 84.7 & \textbf{86.3} & 38.7 & 76.8
      & 0.0327 & 0.0235 & 0.0336 & 0.0299 \\
    \rowcolor{qadrow} \quad QAD $+$ LoRA ($r{=}32$) & \textbf{82.8} & 79.9 & \textbf{90.5} & \textbf{93.7} & \textbf{67.7} & \textbf{86.7} & \textbf{86.3} & \textbf{39.3} & \textbf{78.4}
      & \textbf{0.0281} & 0.0205 & \textbf{0.0293} & \textbf{0.0260} \\
    \rowcolor{qadrow} \quad QAD $+$ LoRA ($r{=}64$) & 79.8 & 79.9 & 87.4 & 93.3 & 65.7 & 84.0 & 85.7 & 36.0 & 76.5
      & 0.0286 & \textbf{0.0204} & 0.0295 & 0.0262 \\
    \bottomrule
  \end{tabular}
  \end{adjustbox}
  \caption{LoRA QAD of Muse-Glimmer 30B, a dense model whose NVFP4
  quantization gap is small. Both adapters are trained for one epoch at
  $5\times10^{-5}$ with an effective batch of 64 sequences of 8192 tokens; the
  RTN baseline uses the same quantized base model without an active adapter.
  Thus, the quantized checkpoints differ only in whether---and how---the LoRA
  correction is trained. \textbf{Bold} marks the best value in each column among the quantized
  checkpoints; the BF16 row is a reference and is excluded.}
  \label{tab:lora_muse}
\end{table}

In Figure~\ref{fig:lora_muse}, forward KL divergence decreases and top-5 Jaccard overlap increases over training, demonstrating that LoRA-QAD improves model fidelity. However, Table~\ref{tab:lora_muse} shows no benefit from increasing the rank: rank 32 outperforms rank 64 in both average accuracy (78.4\% vs.\ 76.5\%) and average KL divergence (0.0260 vs.\ 0.0262). Across both models, increasing the adapter rank does not reliably improve checkpoint quality, and the Qwen3.5 9B results show that it does not close the gap to full-parameter QAD. These results indicate that expanding the low-rank adaptation subspace alone cannot substitute for full-parameter updates.

\subsection{Sequence-Length and Cross-Domain Generalization}
\label{sec:finding-generalization}

\begin{findingbox}
Finding 4: With the same training-token budget, QAD on fewer 32K sequences
generalizes better than QAD on more 4K sequences, improving average accuracy
across AIME25, BigCodeBench, and LiveCodeBench-v6 by 1.9 percentage points. QAD
also generalizes across domains, with the strongest transfer observed from code training to mathematical
reasoning.
\end{findingbox}

\begin{figure*}[!t]
  \centering
  \includegraphics[width=0.8\textwidth]{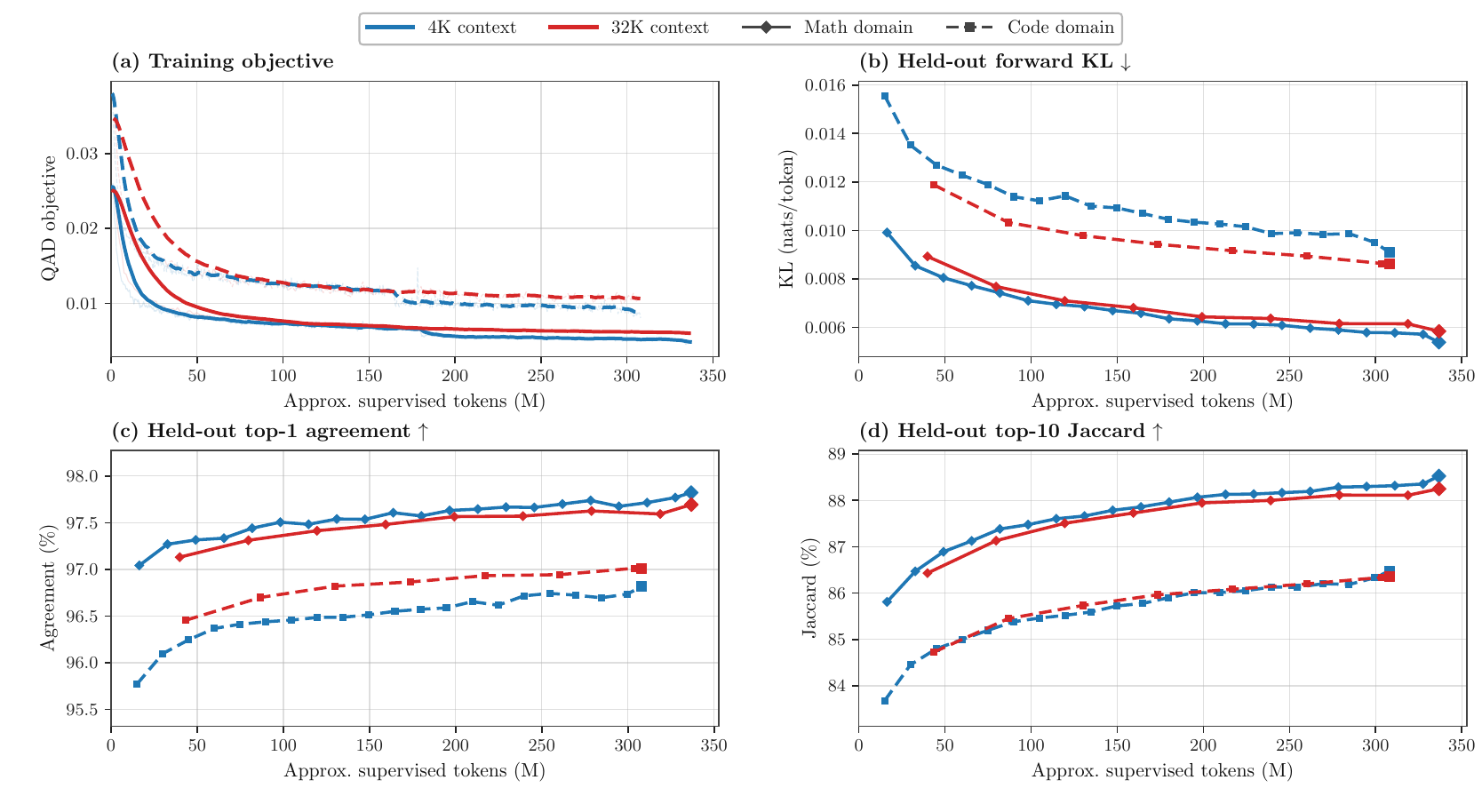}
  \caption{Optimization and held-out fidelity for token-matched Qwen3.5-9B
  NVFP4 QAD. Rows denote the Math and Code source domains. Columns report the
  training objective, forward KL from the BF16 teacher, top-1 agreement, and
  top-10 Jaccard overlap with the teacher. The x-axis reports training progress
  as a fraction of each run's final supervised-token budget; thin training traces are raw logs and thick traces are EMA-smoothed.}
  \label{fig:length_domain_training}
\end{figure*}

We study sequence-length and domain generalization by applying NVFP4 QAD to
Qwen3.5 9B on two 45K-example training corpora, one for mathematics and one for code, each created
by filtering Open Perfect Blend~\citep{perfectblend} for the corresponding
domain. Within each domain,
the 4K and 32K treatments use the same examples with different truncation
lengths. We match the total number of tokens contributing to the supervised training loss by training the 4K treatments
for 5,264 steps and the 32K treatments for 1,817--2,164 steps, yielding
approximately 336.7M tokens for Math and 307.9M tokens for Code. Thus, the 4K
runs process more short sequences, whereas the 32K runs process fewer but
longer sequences. Figure~\ref{fig:length_domain_training} shows that all four
treatments optimize stably and converge toward the BF16 teacher. We then
evaluate the resulting checkpoints on AIME25, BigCodeBench (BCB), and
LiveCodeBench-v6 (LCB-v6), as reported in
Table~\ref{tab:length_domain_generalization}.

\begin{table*}[!t]
  \centering
  \footnotesize
  \setlength{\tabcolsep}{5pt}
  \begin{adjustbox}{max width=\textwidth}
  \begin{tabular}{llrrrrrr}
    \toprule
    \textbf{Inference} & \textbf{Training treatment} & \textbf{Steps}
    & \textbf{Tokens (M)} & \textbf{AIME25} & \textbf{BCB}
    & \textbf{LCB-v6} & \textbf{Avg.} \\
    & & & & $\uparrow$ & $\uparrow$ & $\uparrow$ & $\uparrow$ \\
    \midrule
    BF16 & Teacher (reference) & --- & --- & 54.67 & 30.67 & 86.33 & 57.22 \\
    \addlinespace
    W4A16 & Math-4K  & 5,264 & 336.719 & 47.33 & 28.67 & 79.00 & 51.67 \\
    \rowcolor{qadrow}
    W4A16 & Math-32K & 2,164 & 336.776 & 51.33 & 30.00 & 81.67 & 54.33 \\
    W4A16 & Code-4K  & 5,264 & 307.896 & 46.67 & 30.67 & 79.67 & 52.34 \\
    \rowcolor{qadrow}
    W4A16 & Code-32K & 1,817 & 307.901 & \textbf{52.00} & \textbf{34.00}
      & \textbf{82.67} & \textbf{56.22} \\
    \addlinespace
    W4A4 & Math-4K  & 5,264 & 336.719 & 42.00 & 25.33 & 72.00 & 46.44 \\
    \rowcolor{qadrow}
    W4A4 & Math-32K & 2,164 & 336.776 & 44.00 & 23.67 & 74.67 & 47.45 \\
    W4A4 & Code-4K  & 5,264 & 307.896 & 45.33 & \textbf{26.33} & 77.67
      & \textbf{49.78} \\
    \rowcolor{qadrow}
    W4A4 & Code-32K & 1,817 & 307.901 & \textbf{46.67} & 24.33
      & \textbf{78.33} & \textbf{49.78} \\
    \bottomrule
  \end{tabular}
  \end{adjustbox}
  \caption{Token-matched length--domain generalization. AIME25 is averaged over five samples per problem; BigCodeBench (BCB) and LiveCodeBench-v6 (LCB-v6) use 300 problems each, all with a 32K maximum generation budget. Avg.\ is their unweighted average. Bold marks the best benchmark result within
  each inference dtype; shaded rows are the 32K treatments.}
  \label{tab:length_domain_generalization}
\end{table*}

The results show a clear advantage for longer training sequences. Across the
four pairwise comparisons defined by source domain and inference precision, 32K training improves the three-task
average in three cases and ties in the fourth. Pooled across all comparisons,
average accuracy increases from 50.1\% to 51.9\%, an absolute gain of 1.9
percentage points. The 32K treatment improves AIME25 and LCB-v6 in every
comparison. BCB is the only task that declines with 32K training in both W4A4 comparisons. Therefore, given
the same number of training tokens, using fewer long sequences is generally
more effective than using more short sequences, although the benefit varies by
task.

QAD improvements also extend beyond the training domain. In particular,
Code-trained checkpoints match or outperform their Math-trained counterparts
on AIME25 in three of four matched settings, while outperforming the corresponding
Math-trained checkpoints on both coding benchmarks. Math-trained checkpoints likewise remain effective on both
coding benchmarks. These results indicate that QAD does not merely fit the
source domain: training on one domain can improve or preserve model quality in
another. In our experiments, this transfer is strongest from code to
mathematical reasoning.

\subsection{Quantization-Aware Reinforcement Learning (QARL)}
\label{sec:finding-qarl}

\begin{findingbox}
Finding 5: NVFP4 QARL reduces the rollout bottleneck and accelerates end-to-end
RL training by \(1.23\times\). It also improves mean accuracy across five
mathematical reasoning benchmarks by 2.7 percentage points compared with BF16
RL followed by post-training quantization.
\end{findingbox}

\begin{figure*}[!t]
    \centering
    \includegraphics[width=\textwidth]{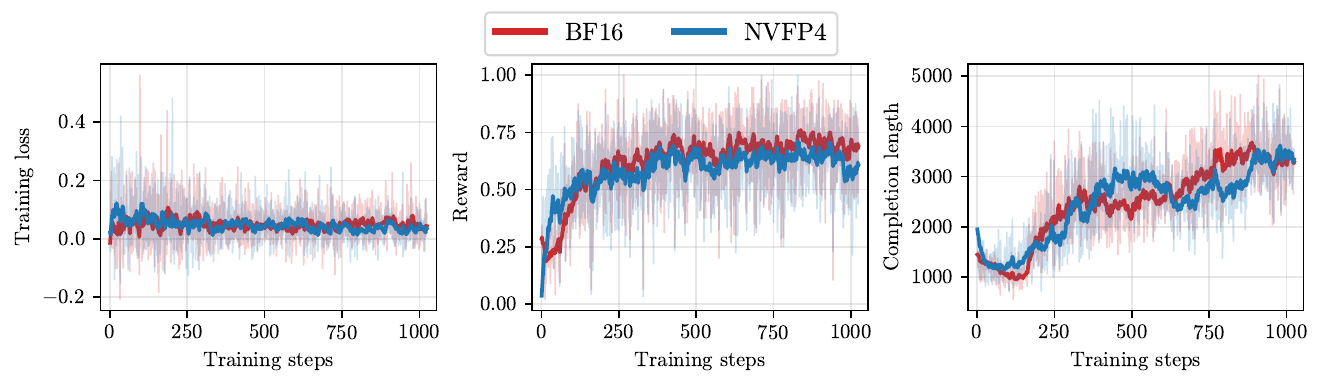}
    \caption{Training loss, rollout reward, and completion length over 1,024 optimizer steps for BF16 GRPO and NVFP4 QARL on Qwen3-8B-Base. NVFP4 rollouts use real quantized weights in vLLM, while the training forward pass uses simulated quantization.}
    \label{fig:qarl-training}
\end{figure*}

Using the setup described above, we compare NVFP4 QARL with BF16 GRPO.
Figure~\ref{fig:qarl-training} shows that NVFP4 QARL remains stable, with loss,
reward, and completion-length trajectories similar to those of BF16 GRPO.
We evaluate checkpoint quality on AIME24~\citep{aime24},
AIME25~\citep{aime25}, MATH500~\citep{math500}, OlympiadBench~\citep{he2024olympiadbench},
and GSM8K~\citep{cobbe2021training}.

\begin{table*}[!t]
    \centering
    \small
    \setlength{\tabcolsep}{5pt}
    \begin{adjustbox}{max width=\textwidth}
    \begin{tabular}{l cc rrrrr r}
      \toprule
      \textbf{Checkpoint} & \textbf{Training precision} & \textbf{Eval precision}
      & \textbf{AIME24} & \textbf{AIME25} & \textbf{MATH500}
      & \textbf{Olympiad} & \textbf{GSM8K} & \textbf{Mean} \\
      \midrule
      Qwen3-8B-Base (untrained) & --- & BF16
        & 5.8 & 1.7 & 35.4 & 18.4 & 43.5 & 21.0 \\
      BF16 RL & BF16 & BF16
        & 20.0 & \textbf{17.1} & \textbf{82.5} & \textbf{49.0} & \textbf{91.9} & \textbf{52.1} \\
      BF16 RL $\rightarrow$ RTN & BF16 & NVFP4
        & 18.3 & 13.8 & 76.3 & 46.3 & 85.6 & 48.1 \\
      \rowcolor{qadrow}
      QARL & NVFP4 & NVFP4
        & \textbf{20.8} & 16.7 & 80.4 & 48.4 & 87.9 & 50.8 \\
      \bottomrule
    \end{tabular}
    \end{adjustbox}
    \caption{Reasoning benchmark accuracy for Qwen3-8B-Base before RL, after BF16 RL, after BF16 RL followed by NVFP4 round-to-nearest (RTN) quantization, and after NVFP4 QARL. Values are percentages; higher is better.}
    \label{tab:qarl-quality}
\end{table*}

QARL improves both quantized-model quality and training speed. As shown in
Table~\ref{tab:qarl-quality}, applying NVFP4 quantization after BF16 RL reduces
mean accuracy across the five reasoning benchmarks from 52.1\% to 48.1\%,
whereas training with NVFP4 QARL reaches 50.8\%. QARL therefore exceeds BF16 RL
followed by post-training quantization by 2.7 percentage points and remains
within 1.3 percentage points of the BF16 RL checkpoint; it also slightly
outperforms BF16 RL on AIME24. Table~\ref{tab:qarl-timing} shows that native
NVFP4 rollouts reduce generation time from 37.46\,s to 28.31\,s per step. This
gain more than offsets the additional quantization and synchronization costs,
reducing total step time from 41.83\,s to 33.96\,s and increasing throughput by
\(1.23\times\). Together, these results show that accounting for quantization
during RL produces a better quantized model than applying quantization
afterward, while also making training faster.

\begin{table*}[!t]
    \centering
    \small
    \setlength{\tabcolsep}{6pt}
    \begin{adjustbox}{max width=\textwidth}
    \begin{tabular}{l rrrrr}
      \toprule
      \textbf{Method}
      & \textbf{Weight sync}
      & \textbf{vLLM rollout}
      & \textbf{Training forward}
      & \textbf{Backward}
      & \textbf{Total} \\
      \midrule
      \rowcolor{qadrow}
      NVFP4 QARL & 0.99 & \textbf{28.31} & 1.52 & 3.14 & \textbf{33.96} \\
      BF16 GRPO & \textbf{0.73} & 37.46 & \textbf{0.98} & \textbf{2.66} & 41.83 \\
      \bottomrule
    \end{tabular}
    \end{adjustbox}
    \caption{Wall-clock time per optimizer step, in seconds, broken down by stage. The backward measurement includes activation-checkpoint recomputation. Lower is better.}
    \label{tab:qarl-timing}
\end{table*}

\subsection{Guidelines for Quantization-Aware Training}
\label{sec:summary-guidelines}

Our experiments suggest four practical guidelines for low-precision training:
\begin{enumerate}
    \item \textbf{Choose activation precision for each quantization format.}
    Deployment precision alone should not determine training precision. For
    NVFP4 W4A4 deployment, W4A16 training generally produces better aggregate
    accuracy and KL divergence, whereas MXFP4 benefits from W4A4 training.
    Practitioners should compare exported checkpoints rather than rely only on
    training-time metrics.

    \item \textbf{Use LoRA to reduce memory, but do not assume that larger ranks
    improve quality.} In our Qwen3.5 9B experiment at an 8K sequence length,
    rank-16 LoRA reduces GPU memory usage by \(2.9\times\). However,
    increasing the adapter rank does not reliably improve accuracy or KL
    divergence and does not close the gap to full-parameter QAD.

    \item \textbf{Prefer longer sequences when the training-token budget is
    fixed.} Training on fewer 32K sequences generalizes better overall than
    training on more 4K sequences with the same number of tokens. QAD updates
    also transfer across domains, with code-domain training improving
    mathematical reasoning in our experiments.

    \item \textbf{Prefer quantization-aware reinforcement learning to post-training quantization for better quality and efficiency.}
    When native low-precision rollout execution is available, combining
    quantized rollouts with quantization-aware updates can improve both speed
    and model quality. In our NVFP4 experiment, QARL increases throughput by
    \(1.23\times\) and improves mean accuracy across five reasoning benchmarks
    by 2.7 percentage points over BF16 RL followed by post-training
    quantization.
\end{enumerate}

\section{Conclusion and Future Directions}
\label{sec:conclusion}

We introduced \projectname, an open, deployment-aligned framework for
quantization-aware training through distillation and reinforcement learning.
By reproducing the quantization behavior of NVFP4, MXFP4, and Q4\_K while
performing computation in BF16, \projectname enables efficient and accessible
QAT without requiring native hardware support for the target format. Its unified
quantization and export abstractions also make the framework extensible across
formats, inference engines, and training methods. \projectname supports dense
and MoE models from 8B to 230B parameters and exports checkpoints directly to
vLLM, SGLang, and \texttt{llama.cpp}. Across these settings, QAD outperforms
strong post-training quantization baselines. Our experiments also show that
effective low-precision training requires format- and workload-specific
choices: activation quantization helps MXFP4 but not NVFP4, LoRA substantially
reduces memory use but does not match full-parameter QAD, longer training
sequences improve generalization under a fixed token budget, and NVFP4 QARL
improves both deployed-model quality and training throughput over BF16
reinforcement learning followed by quantization.

Future work will extend \projectname to additional quantization formats,
inference engines, model architectures, and QAT methods. This includes
lookup-table-based quantization formats planned for NVIDIA's upcoming Rubin
GPUs. More broadly, our goal is to make QAT and QAD accessible to the community
and provide a common, deployment-aligned platform for developing and evaluating
new quantization-aware training techniques.

\bibliographystyle{plainnat}
\bibliography{main}

\newpage
\appendix








\newpage

\end{document}